# Language for Description of Worlds

Dimiter Dobrev
Institute of Mathematics and Informatics
Bulgarian Academy of Sciences
*d@dobrev.com*

We will reduce the task of creating AI to the task of finding an appropriate language for description of the world. This will not be a programing language because programing languages describe only computable functions, while our language will describe a somewhat broader class of functions. Another specificity of this language will be that the description will consist of separate modules. This will enable us look for the description of the world automatically such that we discover it module after module. Our approach to the creation of this new language will be to start with a particular world and write the description of that particular world. The point is that the language which can describe this particular world will be appropriate for describing any world.

## 1. Introduction

This paper presents a new approach to the exploration of AI. This is the Event-Driven (ED) approach. The underlying idea of the ED approach is that instead of absorbing all input/output information, the model should reflect only the events which matter ("important events").

Every action is an event. Every observation is an event, too. If the model were to reflect each and every action and observation, it would end up overloaded with an enormous amount of information. The overloaded situation can however be avoided when the model is limited only to certain important events. This leads us to the idea of Event-Driven models.

A disadvantage (or perhaps an advantage) of the ED model is that it does not describe the world completely, but only partially. More precisely the ED model describes a certain class of worlds (the worlds which comply with a certain pattern).

**What makes this paper different?** The mainstream approach to dealing with multi-agent systems is based on the assumption that the world is given (known) and what we need to find is a policy. In other words, the world is part of the known terms of the problem while the policy is the unknown part. This paper is different because we will assume that instead of being given, the world is unknown and has to be found.

The assumption that the world is given implies that we have a relation which provides a full description of the world. Conversely, in this paper we will try to provide partial descriptions of the world and will do so by employing ED models.

**Structure of the language:** The description of the world will not be similar to a homogenous system consisting of one single layer. Our description of the world would rather be structured as a multilayer system. In Figure 1 we have presented our multilayer structure as a pyramid where the first layer is the base of the pyramid:

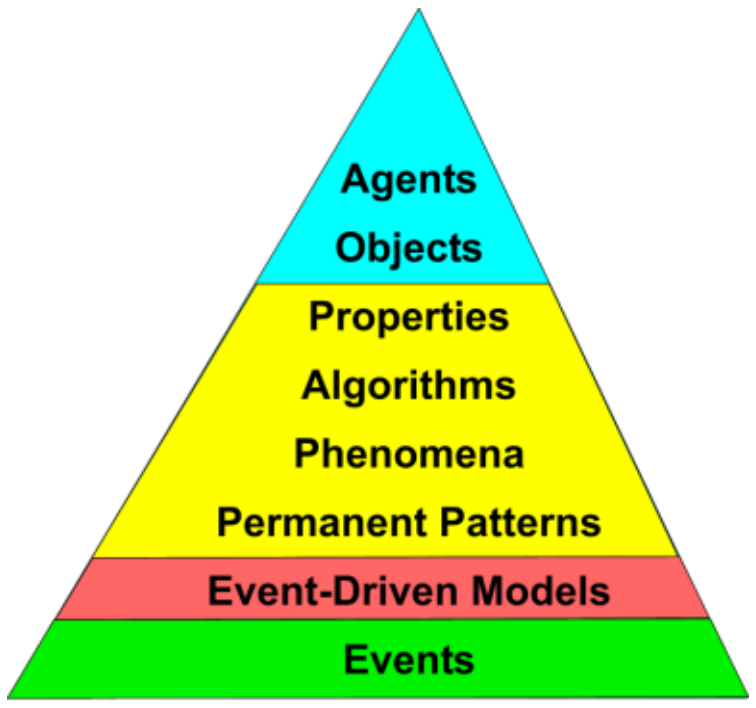

Figure 1

Thus, events will form the first layer of our system. We will use these events in order to describe the Event-Driven (ED) models. Importantly, the states of the ED models must not be the same (some states must be different). For this condition to be fulfilled, there must be at least one state with a special occurrence in it so that the state can be distinguished from other states. We will name this special occurrence *distinction*. The set of the distinctions we will designate as *trace*.

Our first idea about the trace is that it is fixed. (For example, let us have a state in which is always cold.) Then our notion of a trace will evolve and we will have a *moving* trace which can appear and then disappear or move from one state to another. (In our example the coldness could move to the room next door.)

The next layers of the pyramid will consist of various patterns (we will present all of them by ED models). We will begin with the permanent patterns, i.e. those which are observed all the time. While most authors assume that all patterns are permanent, in this paper we reckon that in addition to the permanent ones there other, impermanent patterns which are observed only from time to time. We will name these impermanent patterns *phenomena*. In other words, just as traces can be *fixed* or *moving*, patterns can also be *permanent* or *phenomena*.

Algorithms are also impermanent patterns (observed only when an algorithm is being executed). We will present algorithms as sequences of events. Typically, algorithms are understood as sequences of *actions* on the basis of the assumption that the protagonist is always the one who executes the algorithm. In this paper the algorithm will be any pattern and if the actions of an agent are aimed at maintaining the pattern, then we will say that the agent is the one who executes the algorithm.

When a phenomenon is associated with the observation of an object, we will call that phenomenon a *property*. This takes us to an abstraction of a higher order, namely the abstractive concept of *object*. Objects are not directly observable, but are still identifiable through their properties.

The next abstraction we get to will be *agents*. Similar to objects, we cannot observe agents directly, but can still gauge them on the basis of their actions. In order to describe the world, we have to describe the agents which live that world and explain what we know about these agents. The most important thing to describe about agents is whether they are our friends or foes and accordingly what will they try to do to us by their actions – help us or disrupt us?

The descriptions above relate to computable worlds (ones that can be emulated by a computer program). While any presence of a non-computable agent in the world would make the world itself non-computable, there is another way to make non-computable worlds. We may add a rule which depends on the existence of some algorithm (more precisely, on the existence of an execution of that algorithm). The question “Does an execution of the algorithm exist?” is non-computable (halting problem, Turing (1937)).

**Contributions**

1. Event-Driven model. (The concept has already been introduced by Dobrev (2018), but that paper did not provide an interpretation of the ED model. This is important, because interpretations are what makes models meaningful and distinguish between adequate and inadequate models.)

2. This paper proves that Markov Decision Process (MDP) is a special case of an ED model and that an ED model is the natural generalization of MDP.

3. Simple MDP. We will simplify the MDP to obtain a more straightforward model which can describe more worlds.

4. Extended model. This is the model in which the state knows everything. We will use this model in order to introduce an interpretation of events and ED models. (Although the Extended model was introduced in Dobrev (2019a), in that version of the model the state knows only what has happened and what is going to happen, but does not know what is possible to happen. Therefore, the state of the Extended model in Dobrev (2019a) knows nothing about the missed opportunities. In Dobrev (2019a) the Extended model is referred to as "maximal".)

5. A definition of the concept *algorithm*. We have presented the algorithm as a sequence of events in arbitrary world. Further on, we present the Turing Machine as an ED model found in a special world where an infinite tape exists. Thus we prove that the new definition generalizes the *Turing Machine* concept and expands the *algorithm* concept.

6. A language for description of worlds such that the description can be searched automatically without human intervention.

**How is this paper organized.** First (in Section 2) we will identify the particular world which we are going to describe. Then (in Section 3) we will prove that the known tools for description of worlds are not appropriate for the world in question.

Sections 4 to 8 will provide the theoretical basis of the paper. We will define the concepts *lived experience*, *life*, *Event-Driven model*, *event* and *world*. We will give meaning to these concepts by defining interpretation. We will prove that MDP is a special case of an ED model.

In Sections 9 to 13 we will provide a concrete description of the world in which the agent plays a chess game:

In Section 9 we will describe some simple patterns and will present them by using ED models (such as the patterns Horizontal and Vertical).

Section 10 will define the rules to which the chess pieces move. For this purpose we will need to expand the *algorithm* concept. The algorithms to which chess pieces move will also be presented through ED models.

In Section 11 we will present the chess pieces as objects. The objects will be abstraction of higher order. They will be defined through properties. The property is something concrete and it will be defined through an ED model.

In Section 12 we will add a second player. For this purpose we will make another higher-order abstraction. This will be the abstraction *agent*. We will not observe agents directly and instead will gauge them through their actions. Agents will make the world non-computable, but if we add some non-computable rule then the world can become non-computable even without agents.

Finally, in Section 13 we will look at the agents and various interplays between them.

## 2. The chess game

Which concrete world are we going to use in order to create the new language for description of worlds? This will be the world of chess.

Let us first note that we will want the world to be partially observable because if the agent can see everything the world will not be interesting. If the agent sees everything, she will not need any imagination. The most important trait of the agent is the ability to imagine the part of the world she does not see at the current moment.

For the world to be partially observable we will assume that the agent sees just one square of the chessboard rather than the entire board (Figure 2). The agent's eye will be positioned in the square she can see at the moment, and the agent will be able to move that eye from one square to another so as to monitor the whole board. Formally speaking, there is not any difference between seeing the whole board at once and exploring it by checking one square at a time – in either case one gets the full picture. There will not be any difference only if you know that by moving your sight from one square to another you will monitor the whole chessboard. In practice the agent does not know anything, so she will need to conjure up the whole board, which however will not be an easy process and will require some degree of imagination.

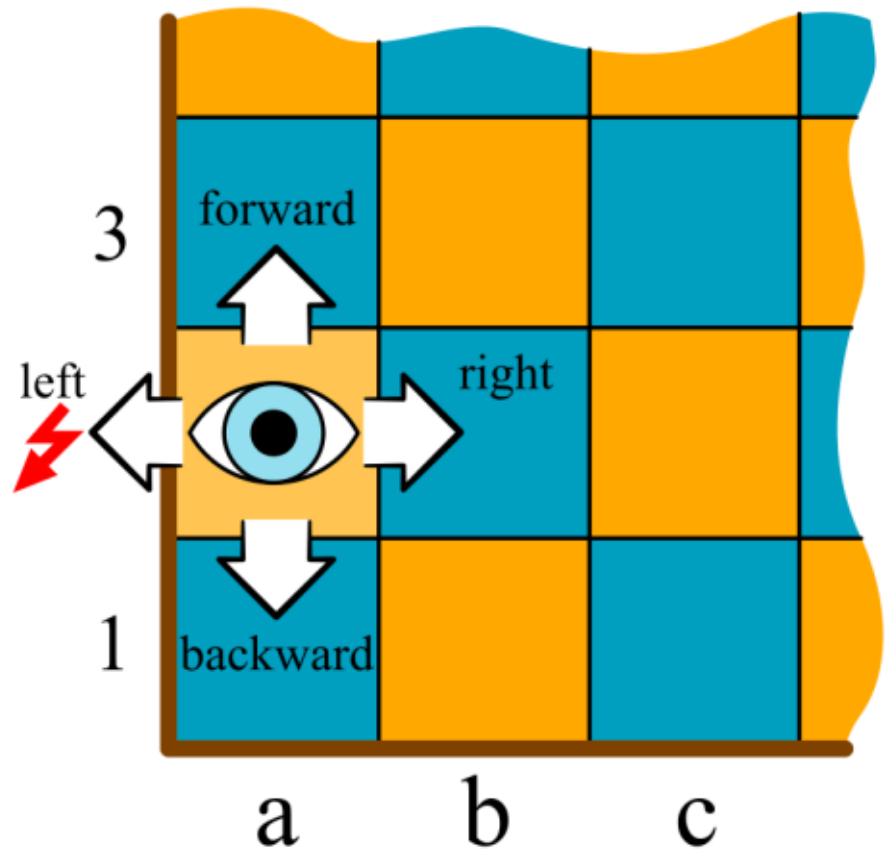

Figure 2

In Figure 2, the agent has her eye in square **a2** and can move it in all the four directions. (Right now she cannot move it left because this is the edge of the board and a move to the left would be incorrect.) In addition to moving the eye in the four directions, the agent can perform two other actions: *"Lift the piece you see right now"* and *"Put the piece you lifted in the square you see right now"*. We will designate the two additional actions as *Up* and *Down*. These six actions will enable the agent monitor the chessboard and move the chess pieces, and that's everything one needs to play chess.

### 2.1 A chess game with a single player

We will examine two versions of a chess game – a game with a single player and a game with two players.

How do you play chess with a single player? You first move some white piece, then turn the board around, play some black piece and so forth.

We will start by describing the more simple version in which the agent plays against herself. This version is simpler because in that world there is only one agent and that agent is the protagonist. Next we will examine the more complicated version wherein there is a second agent in the world and that second agent is an opponent of the protagonist.

The question is what can be the goal when we play against ourselves?

## 2.2 The goal

While the authors of most papers dedicated to AI choose a goal, in this paper we will not set a particular goal. All we want is to describe the world, and when we figure out how the world works we will be able to set various goals. In chess for example our goal can be to win the game or lose it. When we play against ourselves, the goal can vary. When we play from the side of the white pieces our goal can be *"white to win"* and vice versa.

Understanding the world does not hinge on the setting of a particular goal. The Natural Intelligence (the human being) usually does not have a clearly defined goal, but that does not prevent him from living.

There are two questions: "What's going on?" and "What should I do?" Most authors of AI papers rush to answer the second question before they have answered the first one. In other words, they are looking for some policy, and a policy can only exist when there is a goal to be pursued by that policy. We will try to answer only the first question and will not deal with the second one at all. Hence, for the purposes of the present paper we will not need a goal.

In most papers the goal is defined through *rewards* (the goal is to collect as many rewards as possible). When referring to Markov decision process (MDP) we will assume that rewards have been deleted from the definition because we need them only when we intend to look for a policy, while this paper is not about finding policies.

## 3. Related work

Can the chess world be described using the already known tools for description of worlds? We will review the tools known at this time and will prove that they are not appropriate.

## 3.1 Markov decision process

The most widely used tool for description of worlds is Markov decision process (MDP). Can we use MDP for describing the kind of world selected in this paper? Let us first note that we will have to use Partially observable MDP (POMDP) because the world we are aiming to describe is Partially observable.

Certainly, the chess world can be presented as a POMDP, but how many states will this take? We will need as many states as the positions on the chessboard are, which is an awful lot (in the range of $10^{45}$ according to Tromp (2021)). We will even need some more states because in addition to the chessboard position the state must remember the whereabouts of the eye. This means 64 times more states, which is a little more because adding two zeroes to a large number seems an insignificant increase. Thus, we do not perceive the numbers $10^{45}$ and $10^{47}$ as much different.

If we wish to describe this kind of POMDP in tabular form, the description will be so huge that storing it would be beyond the capacity of any computer memory. Of course storing the description is the least problem. A much more serious issue is that we should find and build that table on the basis of our lived experience which means that for such a huge table we would need enormously huge lived experience (almost infinite).

Therefore, efforts to find a POMDP description of the chess are bound to fail.

## 3.2 Situation Calculus

The first language for description of worlds was proposed by Raymond Reiter and this is *Situation Calculus* described in Reiter (2001).

The chess world can indeed be presented through the formalism proposed by Raymond Reiter, but we will run into two problems (a small one and big one).

The first (smaller) problem is that in order to present the state of the world obtained after a certain action, Reiter uses a functional symbol. Thus, Reiter assumes that the next state of the world is unambiguously defined. That assumption would not be a problem for deterministic worlds such as the chess game, but would be an issue when it comes to a dice game. The problem of course is not a big one because we can always assume that the next state is determinate even if we do not know which one it is. (In other words, we can assume that there is some destiny which defines the future in an unambiguous manner, although it does not help us predict what is going to happen.)

In Section 3.2. of their publication of 2001, Boutilier, Reiter and Price (2001) attempt to resolve the first (smaller) problem by replacing each step with two steps (plies). Instead of a single step of the agent they suggest two plies – *agent ply* and *nature ply*. The idea is that the agent step is non-deterministic because the world (nature) can respond in a variety of ways, so if we decouple the agent's action from nature's response then the output from each ply would be deterministic. Essentially we use the same idea by creating a Simple MDP (further down in this paper).

The second (big) issue with Situation Calculus is Reiter's implied assumption that there is some human being (programmer) who has figured out how the world is designed and will describe that design using first order formulas. All of us wish to get to a description of the world by first order formulas, but the objective is to ensure that the description can be found automatically, i.e. without human intervention. While this paper actually provides a manmade description of the chess game, our objective is to come up with a machine-searchable description and the one provided here can indeed be searched and found automatically.

## 4. Lived experience

The task is to describe the world on the basis of our lived experience. This means we are looking for a model which provides the best explanation of what has happened until the present moment. Hence the first question we will address is "What is lived experience?".

The common assumption is that we have only one life and lived experience comprises everything that has happened in that life until now. However, in this paper we will assume that we have more than one life and lived experience comprises everything that has happened both in the current life and in all previously lived lives. We will go even further and assume that we may have more than one current life.

Why do we make this assumption? From the individual's perspective each individual has only one life, but from a population perspective there are many lives. In our case we have an artificial agent living in an artificial world. We can let the agent live many times in the world and will thus obtain many lives from which we can collect observations and try to identify patterns. We can even let multiple agents live in the world concurrently. Then will have multiple current lives.

**Definition:** Lived experience is a finite sequence of lives.

This takes us to the next question: "What is life?"

### 4.1 Life

*Life* will be a finite sequence of actions and observations. Life is what we see, but are there things in life which we do not see?

Let us imagine we have had two lives. In the first life we walked past a pot full of gold coins, but did not bother to open it and went on. In the second life we walked past an empty pot which we did not open either and again went on. In terms of actions and observations the two lives are fully identical, but they are still two different lives because in the first one we had the chance to find plenty of gold coins while in the second life we never had that chance.

That is why our description of life will consist of two determinants. The first one is what we have seen (the sequence of actions and observations). We will label this sequence as the *trace of life*.

**Note:** In this paper we use the terms *trace of live* and *trace of the ED model*. These are different things.

The second determinant of life is everything that was possible to happen in that life (regardless of whether it actually happened). The possible past and the possible future are described by the state of the world. Knowing what the state has been at each moment of time is important. That is why to the description of life we will add the sequence of states which the world has been through. We will label this sequence as the *backbone of life*.

The trace and the backbone describe life from the perspective of the world. If we look at the world from the agent's perspective, life will be described by the trace and by the guesses which the agent can make at any given moment. For example, at a given moment we can assume that after two steps we will see a certain observation. (The moment of the guess is the moment at which we are able to make that guess rather than the moment which the guess relates to).

Two things are important for each guess. First, the antecedents which must be true at the moment when we make the guess. The second one is the rule (heuristic) on the basis of which the guess can be made. We will assume that heuristics are derived from our entire lived experience (including the future lived experience). Hence, the addition of more lived experience leads to a change of heuristics and thereby to a change of guesses (even previous guesses). For example, there comes a moment when we realize that the sound produced by a fire weapon signifies some danger. Then we review our past and find out every time we heard a fire shot we had been in trouble.

## 4.2 Heuristic

We said that in life there are things we do not see but still important enough for us to try predicting them. For example, "Are there gold coins in the pot?" We may try to predict this through our intuition or on the basis of some heuristics which we derive by collecting statistics from our lived experience. An example of such heuristic is "If the pot looks ancient, then it is full of gold coins". Certainly, each heuristic comes with some confidence coefficient. Let this coefficient be as low as 2%. Even with such a low confidence coefficient, if it appears to be an ancient pot it is worth the effort to open it and check for gold coins.

An heuristic will consist of antecedent, consequent and confidence coefficient. We will suppose that the consequent is an atomic formula, while the antecedent is a conjunction of atomic formulas. (We will define atomic formulas later.)

**Definition:** An heuristic is a material implication with confidence coefficient

$$A_1 \,\&\, A_2 \,\&\, \dots \,\&\, A_n \Rightarrow A_0 \text{ (per cent)}$$

Here is an example of a heuristic:

$$Ob(t\text{–}1)=o_1 \,\&\, Act(t)=a \Rightarrow Ob(t+1)=o_2 \text{ ( 10\% )}$$

Here $Ob(t)=o$ means that the observation at moment $t$ is $o$ and $Act(t)=a$ means that the action at moment $t$ is $a$. (We have expressed this as an equality because the action and the observation depend unambiguously on moment $t$.)

The event $Ob(t)$ makes sense only at even-number moments, while $Act(t)$ makes sense only at odd-number moments because we alternate actions and observations. The heuristic is valid for any $t$. That is why we will assume that the antecedent is known at moment $t$ and this is the first moment in which the antecedent becomes known. (We can achieve this by replacing $t$ with $t+i$ for some $i$.)

The consequent here relates to a future visible event (predicting past visible events does not make sense). The consequent might also be related to an invisible events (past or future). For example:

$$Ob(t)=o_1 \Rightarrow PossNext(o_2, t-2) \text{ (some per cent)}$$

Here the invisible event $PossNext(o, t)$ means it is possible that the next observation is $o$, i.e. it is possible that $Ob(t+2)=o$. (In this expression we have not used equality because the possible next observations are more than one, while an equality implies one possibility.)

For $o_1=o_2$ the above heuristic will be meaningless because it will be tautologically true, while $o_1 \neq o_2$ will make sense for any past or future moment.

We will consider heuristics as rules which are always valid, but might also assume that a rule is valid only sometimes. In this case one impermanent rule will define one invisible event. That event will be True when the rule is valid and False when the rule is invalid. Let us consider the following rule:

$$\varnothing \Rightarrow Ob(t+2) \neq o$$

The antecedent here is the empty set, i.e. without antecedent or always. This rule tells us that the next observation cannot be $o$. This is sometimes True and sometimes False, but we do not see whether this is True (unless the next observation is $o$, but generally we cannot see that). Therefore, this is one invisible event and its negation is exactly $PossNext(o, t)$.

## 4.3 Atomic formula

The agent cannot observe all events and will have to stay with some finite number of events. The agent will assign a name to each observed event and we will call these events atomic formulas.

**Definition:** An atomic formula is an event to which we have assigned a name.

For example, $Ob(t)=o$ is an atomic formula. (More precisely, these are multiple atomic formulas, one for each $o$.)

When observing some event we will actually be observing its negation as well. Thus, the negation of an atomic formula is also an atomic formula.

A conjunction of atomic formulas can also be an atomic formula if we have decided to observe it and provided that we have assigned a name to it. These conjunctions of course will be infinitely many and we would not be able to observe them all.

From event $A(t)$ we can make countably many events $A(t+i)$, which will be the same event shifted in time. It does not make much sense to observe all these events because they are almost the same. Therefore we will observe only one of them. For example, if we have a conjunction of atomic formulas, we will decide to observe the event at the moment which is the maximum of the moments of the atomic formulas included in the conjunction.

In most cases the moment at which an observed event occurs is not important, but there are some exceptions. With ED models for example the exact time at which each of the observed events occurs is important (it is important whether an event occurs several steps earlier or several steps later).

## 4.4 Single-moment event

**Definition:** A single-moment event is one which depends only on one moment.

Let that moment be $t$. Then, a visible single-moment event is an event which depends only on one observation or only on one action (only on $Ob(t)$ or only on $Act(t)$). An invisible single-moment event is one which depends only on one state of the world ($s_t$).

A conjunction of single-moment events can also be a single-moment event if all atoms in the conjunction are single-moment events and relate to one and the same moment.

The visible single-moment events are finitely many and there are as many invisible single-moment events as are the subsets of the states of the world (finite or continuum).

An atomic formula may or may not be a single-moment event.

### 4.5 The guess

At certain moments of time the agent will apply heuristics in order to derive some guesses. Each guess will consist of some consequent and some confidence coefficient. The consequent will be an atomic formula and that formula will relate to a concrete moment of time *t* (which can be in the past or in the future). Hence, unlike the heuristic which applies to any *t*, the guess will apply to a particular *t*.

**Definition:** A guess is expressed as follows:

$$A(t)\ (per\ cent)$$

The set of guesses (heuristic applications) will be called *Guesses*.

Let us have another heuristic: "If the pot is full of gold coins, then we can buy an aircraft". Let the confidence coefficient of this heuristic be 50%. Then as soon as we see a pot which appears ancient we apply the first heuristic and get "the pot is full of gold coins" (with a confidence coefficient of 2%). Now we can buy an aircraft (with a confidence coefficient of 1%).

The important takeaway from this example is that heuristics can be cascaded and that heuristic antecedents may include an invisible event. The cascade application of two heuristics can be considered as new one. However, it will take a lot of lived experience to find this new heuristic. The better way is to find more simple heuristics and combine them in more complex ones. We can find many heuristics which indicate that the pot is full of gold coins and their cascade application will lead to the consequent that in all these cases we can buy an aircraft. If we do not use cascade application then for every particular case we should collect individual statistics which will prove that in that particular case we can buy an aircraft.

### 4.6 The definition of life

In this definition life will be presented as a game between two players. The first player will be the agent (the protagonist), the second player will be "nature" and it will be the world itself. The agent called "nature" is not free to do whatever she likes. She operates to some rules, although she may have some freedom of choice (free will). Further down we will try to obtain an explanation of the world and will replace "nature" with a group of agents. That group will consist of zero, one, two or more agents. When dealing with the empty group we will reckon that in the world there are not any agents except the protagonist.

Life will consist of two things: trace and backbone.

**Definition:** Life is:

$a_1,\ o_2,\ a_3,\ o_4,\ \ldots,\ a_{k-1},\ o_k$, where $a_i \in \Sigma$, $o_i \in \Omega$.

$u_0,\ w_1,\ u_2,\ w_3,\ \ldots,\ w_{k-1},\ u_k$, where $u_i \in U$, $w_i \in W$.

Here $\Sigma$ are the possible actions, $\Omega$ are the possible observations of the agent, $U$ are the states of the world when it is the agent's turn to make a move and $W$ are the states of the world when it is nature's turn to make a move. The symbol $k$ stands for the length of life (if life is $L$, then its length is $|L|$).

If write the trace and the backbone in one line we get this:

$u_0, a_1, w_1, o_2, u_2, a_3, w_3, \ldots, a_{k-1}, w_{k-1}, o_k, u_k$

From the perspective of the world life consists of a trace and a backbone, from the agent's perspective life consists of a trace and guesses and from their joint perspective life is life with guesses.

**Definition:** Life with guesses:

- trace
- backbone
- agent's guesses (heuristic applications):

$$g_0, g_1, g_2, \dots, g_k, \text{ where } g_i \subseteq Guesses.$$

The set $g_i$ consists of the guesses which the agent can make at moment $i$ (not earlier, but exactly at that moment). All valid guesses at moment $t$ are:

$$G_t = \bigcup_{i \leq t} g_i$$

In some cases the set of guesses $G_t$ may be contradictory. The guesses for some $j$ can be both $A(j)$ and $\neg A(j)$ (perhaps with different confidence). The guesses $A(j)$ and $\neg A(j)$ may occur at the same moment (in one $g_i$), but can also occur at two different moments.

The agent exercises thinking only when it is her turn to make a move, therefore only at even-number moments. If $t$ is an even-number moment, the agent cannot choose the guesses $g_t$ because they are defined by the past, but she can choose the guesses $g_{t+1}$ because they are defined by the past plus the agent's next action. (In other words, the agent will run all possible actions and will choose the one which gives her the best $g_{t+1}$.)

## 5. Event-Driven models

Now we will sort of jump the gun and say what is an Event-Driven (ED) model before we even say what is an event *per se*. Roughly, an ED model is a directed graph where the nodes represent the states of the world and the arrows are labeled with events. We assume that the world keeps staying in the same state until some observed event occurs. Then the world shifts to another state by making a move over the arrow labeled with the event which has just occurred.

Distinctions are important elements of the ED model. A distinction in our language will mean that in a certain state a given event occurs with a probability different from the expected (average) probability. Distinctions will help us in two ways: i) understand in which state of the world we are in right now and ii) predict what is going to happen and what has happened. By *trace* we will mean the set of distinctions.

All arrows and distinctions in the ED model will have some probability. Instead of exact probabilities we will use probability intervals. First, we will explain why we prefer to use probability intervals. Second, we will explain why we make a distinction between things which are impossible and things which are possible but improbable. Third, we will introduce the statistics which will give meaning to these probabilities. And only then we will be able to provide the definition of ED models.

### 5.1 The probability interval

In basketball, when you through the ball at the hoop, what is the probability of success? We may try 100 shots and calculate that probability by statistics. Other factors may also have some impact on that probability, such as the light conditions in the basketball hall or how fatigued you are. We may construct a more complex model which captures those factors as well, but some factors will always fall through the cracks and remain unaccounted for. These unaccounted factors will induce a degree of volatility meaning that the probability will not be an exact value but some interval.

The most important one of these additional factors is our freedom of choice (free will). When we shoot the ball, it matters a lot whether we want or do not want to score a goal. Let us assume that when we do not want the probability of scoring a goal by error is $a$ and when we really want the probability of scoring (by intent) is $b$. Thus, each time we shoot at the hoop the probability of scoring a goal will be in the interval $[a, b]$.

### 5.2 Possible but improbable

Can there be some difference between the impossible and the improbable? We will make a distinction between a missing arrow and an arrow with a probability of zero (i.e. the interval $[0, 0]$).

The probability of a given event may tend to zero. This for example is the case when an event occurs once in ten instances, then once in the next one hundred instances, etc. This kind of event is possible, but its probability is zero (or tends to zero).

When looking at the actions of the world (the observations of the agent) the difference between the impossible and the improbable is minor, but when looking at the agent's actions the difference becomes essential because this is the way to define the concept *incorrect move*.

When the arrows are labeled with actions and when an arrow is missing, then we will assume that the corresponding action is incorrect move. This will be a separate event which we will designate as *semi-visible*. An arrow with a probability of zero will mean that the action is correct but for some reason our agent will never (or almost never) execute that action.

### 5.3 Statistics

Statistics is the main tool which the agent uses in order to understand the world. Here we will deal with two types of statistics. The first one is the *statistics of the agent* and is based only on actions and observations. These are the real statistics available to the agent.

We will examine the second type of statistics. We will name them *statistics of the world*. In addition to actions and observations, these statistics capture the states of the world. We can imagine them as statistics created by some super agent who is capable of monitoring the state of the world. Certainly such a super agent does not exist. We assume that the world knows the state it is in. That is why the world will be able to create these statistics but will not be willing to share them with the agent.

Thus, the statistics of the world are something abstract which the agent cannot see. Nevertheless, we will use these statistics in order to define and give meaning to some of the introduced concepts. For example, what is the probability of moving from one state to another over a given arrow?

**Note:** We will use statistics of the world to determine the probabilities of arrows in the world and in the ED models. The problem is that we may have two different ED models that differ only in their probabilities. We assume that the world has chosen one of these different ED models and determines its probabilities based on statistics.

Let us assume that the probabilities are fixed and we will calculate these fixed probabilities using the statistics of the world. We will make $N$ steps and will count how many times we have been through each state and each arrow (for the case of ED models: how many times we have arrived at the state regardless of the number steps we spent in that state). If we have been $k$ times through a given arrow and $m$ times through the tail (the state from which this arrow departs), then the probability of going through the arrow will be approximately $k/m$. Certainly, when the $N$ value is low the error will be large, but when the $N$ value is high we can assume that the probability is exactly $k/m$. Thus, the probability will be the limit value of $k/m$ when $N$ tends to infinity.

This is how we used the statistics of the world in order to give meaning to the probability of the arrows. Can we also use these statistics in order to define the probability intervals? This will be a bit more difficult. Let us again make $N$ steps. Let $n$ be the largest integer for which $n^2 \leq N$. We will break the interval $N$ into $n$ intervals such that in each interval there are at least $n$ steps. We will derive statistics for each of these $n$ intervals and will obtain some probability $p_i$. The probability interval of the arrow will be $[min(p_i), max(p_i)]$. When $N$ tends to infinity we should obtain the exact value, but for this to be true we will make one assumption.

The assumption will be that the world and the agent change their fixed policies but do that only seldom, therefore any changes of the fixed policy of the ED model are also seldom (see below what is a fixed policy). When the world and the agent execute a fixed policy, then the probability of all arrows in the ED model will also be fixed. If the world and the agent change their policies too often, then this statistical method will not give us the result we need (i.e. will produce intervals which are shorter than the actual ones).

### 5.4 The ED model definition

**Definition:** An Event-Driven model consists of three parts:

Topology:

- $S$ – the set of the states of the model (finite or countable, i.e. most countable).
- $E$ – the set of the observed events (a limited number of events).
- $R \subseteq S{\times}E{\times}S$ – relation between the states, labeled with the observed events.
- Collision reconciliation rule.

Partial trace:

- $Trace \subseteq Distinctions$ – the set of distinctions.

Probabilities:

- Probability: $R \rightarrow [0, 1] \times [0, 1]$ – a function, which at each arrow returns a probability interval.
- Backward Probability: Same as Probability, but related to the past (when we move against the direction of the arrow).

**Definition:** Distinction is a 3-tuple consisting of state, event and probability. The probability is either a probability interval or the constant *never*. The probability must be different from the average probability of the event (the expected value). The set of these 3-tuples is *Distinctions*.

We will assume that an arrow is missing then and only then when a distinction in the trace says that a given event cannot occur in a given state. Thus:

$$\forall s_1 \in S\ \forall e \in E\ (\neg \exists s_2 \in S <s_1, e, s_2> \in R \Leftrightarrow <s_1, e, never> \in Trace)$$

The observed events are those which sit over the arrows of the model. The events involved in the trace will be referred to as *additional events*. There may or may not be some intersection between the observed and the additional events.

A *trace* in our discussion is a *partial trace* because we assume that it contains only part of the distinctions of the model. If we assume that the trace holds all distinctions of the model, then it would be a full trace. (If the model has more than one interpretation then we should be dealing with the trace of the interpretation rather than with the trace of the model.)

The collision reconciliation rule tells us which is the next state of the ED model in case that events $a$ and $b$ of $E$ occur at the same time.

The rule can be deterministic:

1. Event *a* has priority over event *b* and in this case the executed event will be *a*.
2. In this case the two events will be executed sequentially – first *a* and then *b*.
3. A separate arrow tells us where to go if *a* & *b* occurs.

The rule can also be non-deterministic:

4. We select *a* with a probability of *p* (or with a probability within the interval $[p_1, p_2]$).

### 5.5 Interpretation

When is the ED model a model of the world? To answer this question, we will define the concepts *interpretation* and *characteristic*. For this definition we will use the statistics of the world. In other words, we will use the information about the state of the world – something which the agent does not know.

For the ED model to be a model of the world there must be a link between life and the state of the model. We will term this link an *interpretation* which, for any moment of life, will tell us (exactly or approximately) the state in which the ED model is. Hence, *interpretation* is a function which for each moment of life returns a state of the ED model (or a set of states or some *belief*).

An interpretation may not be unique, but if we expand the partial trace it will become unique. (We may expand the trace a little bit or expand it all the way to a full trace.)

Of course, interpretation is not an arbitrary function. It must be consistent with the ED model. When none of the observed events occurs in several consecutive moments, the interpretation function must return one and the same state for all these moments. If one of the observed events occurs, in the next moment the interpretation function must return a new state, provided that in the ED model there is an arrow over this event leading from the previous state to the new state.

We will consider several types of interpretations:

1. Simple interpretation. This is a function which for each state of the world returns a state of the ED model.

2. Unambiguous interpretation. This is a function which for each moment of an infinite life returns a state of the ED model. Each simple interpretation is unambiguous, but not vice versa.

3. Set-based interpretation. Same as the unambiguous interpretation, but returns a set of states of the ED model instead of a single set.

4. *Belief*-based interpretation. Same as the set-based interpretation, but returns some *belief* of states of the ED model (i.e. a set with probabilities).

In order to simplify the discussion we will not deal with interpretations type 3 and 4. So, when we refer to an interpretation, we will mean an unambiguous interpretation.

The functions over the moments of life will be referred to as *multi-value events*. The usual events are two-value – they either occur or do not occur. If an event is *n*-value, we will present it as *n* two-value events which do not intersect (cannot occur at the same time). The interpretation is therefore a multi-value event. We will assume that the interpretation is defined for any infinite life. With finite lives we may have a problem defining some events. If we do not have information about the past or future of a certain life, the value of the event can be "I do not know". For this reason we assume that interpretations are defined over infinite lives.

**Definition:** An infinite life consists of an infinite trace and an infinite backbone (infinity is in both directions):

$$\ldots, o_{-2},\ a_{-1},\ o_0,\ a_1,\ o_2,\ a_3,\ \ldots$$

$$\ldots, u_{-2},\ w_{-1},\ u_0,\ w_1,\ u_2,\ w_3,\ \ldots$$

In one line it looks like this:

$$\ldots, o_{-2}, u_{-2}, a_{-1}, w_{-1}, o_0, u_0, a_1, w_1, o_2, u_2, a_3, w_3, \ldots$$

An interpretation described by visible events will be referred to as a visible interpretation. Each visible interpretation is unambiguous.

We assume that the ED model is described by its topology and its partial trace. On the basis of this description we look for an interpretation (one of the possible interpretations). Then, from the interpretation we can obtain the full trace and the probabilities.

## 5.6 Characteristic

Thus, an ED model must have an interpretation in order to be a model of the world, but this is not enough. That interpretation must have some meaningful characteristic.

Each event has its characteristic – this is a fuzzy set of the states of the world in which the event occurs.

**Definition:** The characteristic of an event is a function which for each state of the world gives the probability of that event to occur in that state.

The difference between *belief* and characteristic is that *belief* describes a single state while a characteristic describes set of states. The sum of the probabilities in *belief* is one, while in the case of a characteristic that sum is between zero and infinity.

An interpretation is a multi-value event, i.e. it consists of multiple events, each one with its own characteristic. The characteristic of an interpretation will be the set of all those characteristics. For the characteristic of an interpretation to be meaningful, all these characteristics must not be the same.

**Note:** We will assume that everything which the states of the world "remember" and "know" is important. (This is the assumption that the model of the world is minimal. Later on we will add this assumption.) Therefore, if the characteristics of two events are different there is something important which makes that difference.

**Note:** When the characteristic of an ED model is not meaningful, then the full trace is the empty set. That is, nothing interesting is going on.

The best case is when the characteristics are sets (rather than fuzzy sets). In this case we will have a simple interpretation and a meaningful characteristic.

It is important to know whether the model "remembers" important things. If it remembers only important things, then we will have a simple interpretation. If the model remembers only unimportant things, then we will not have an interpretation or will end up with an interpretation with a meaningless characteristic.

For example, let us take a Fully observable MDP (FOMDP). This ED model "remembers" the last observation. Let us assume that in our world this is something unimportant. The FOMDP interpretation is visible, but the characteristic of this unique interpretation will be meaningless because whatever the last observation is, the expected state of the world will be one and the same (due to our assumption that in our world this information is unimportant). If in our world the last observation carries some important information, then the characteristic of the FOMDP interpretation would by meaningful.

Let us now assume that the last observation is remembered by the state of the world (i.e. this is important information and each symbol in this information is important). Now that the world "remembers" the last observation, we can sort the states on the basis of this observation and obtain an equivalence relation. Then the FOMDP has a simple interpretation and it coincides with the visible one.

**Note:** The above description applies to permanent patterns with a permanent trace. The pattern can also be transient (phenomenon) and in that case the interpretation will not apply to the full life, but only to certain parts of that life. (Phenomena do not occur all the time but only from time to time.) The moving trace also complicates the model.

### 5.7 Examples

Let us have event *a* and the ED model which remembers whether the occurrences of *a* add up to an even or odd number (Figure 3).

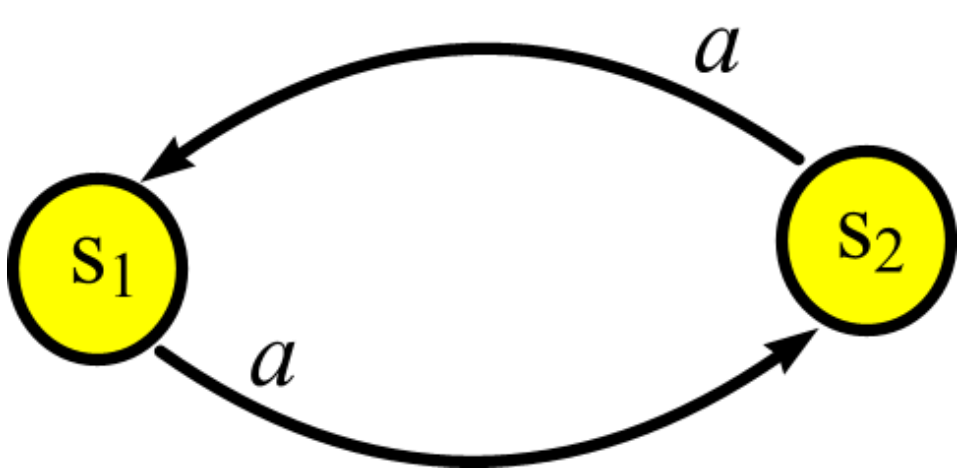

Figure 3

Let us then assume that the odd or even number of occurrences of *a* is something important in our world. Then each state of the world "knows" this and we can sort the events in two sets (a set of even-number occurrences of *a* and all the rest). This gives us two simple interpretations of the ED model. We can relate the set of states in which the count of the occurrences of *a* is an even number to $s_1$ or to $s_2$. Each of these two interpretations will give us a trace. We can take the full trace, but to distinguish the two interpretations it would be sufficient to take a random, non-empty part of the full trace. The arrow probabilities are also determined by the interpretation, but in this case all these probabilities are 1.

Now let us assume that the odd or even number of occurrences of *a* is something completely unimportant in our world. In this assumption the model in Figure 3 remembers only unimportant things. In this case the model will not have any interpretation because the interpretation needs something to start with. Neither does the state of the world know the state of model ($s_1$ or $s_2$), nor can we derive this information from the past and the future. Thus, there is not any function out there which can unambiguously tell between $s_1$ and $s_2$.

If we waive the requirement that the model of the world remembers only important things, we may obtain a new model which will provide some interpretation, however, that interpretation will tell us nothing (will not give us a trace). Let us expand the model of the world by adding that unimportant information (whether *a* occurred an even number of times). We will replace each state *s* with the states <*s*, *even*> and <*s*, *odd*>. If state *s* "remembers" that the count of *a* occurrences is an even number, then state <*s*, *even*> will be reachable, while state <*s*, *odd*> will be unreachable (as if it does not exist). If none of the states remembers that fact, then all these new states will be reachable. In this case $s_1$ will relate to all states <*s*, *even*> or to all states <*s*, *odd*>. In both cases there will be no trace because the probability of each event will be the average probability.

For example, let us imagine our world in which we have the event "new day". Sure the event is important, but whether it is an even-number day or an odd-number day is not important at all. What matters in this case is which day it is over a module of seven because this is the day of the week. The state of our world "remembers" which day it is over a module of seven. This gives us an ED model with seven states which has a trace. The trace is that Sunday is a non-working day. Thanks to this trace we can synchronize any life which is long enough and say, for any moment of time, which day of the week it is (which state the ED model is in). Although the ED model has seven possible interpretations, the partial trace "Sunday is a non-working day" helps us capture one of these possible interpretations.

Another example in which the model has more than one interpretation: Let us have an ED model in which there is some non-deterministic transition to states $s_1$ and $s_2$. Let us split the set of states of the world which correspond to $s_1$ or $s_2$ in two groups on the basis of various properties. We can group them in "blue" and "green" or in "big" and "small". A simple interpretation will correspond to each splitting. These interpretations can be many (if the set is countable the subsets will be a continuum). Certainly, the more distinctions are there in the partial trace, the less interpretations will be possible, while for a full trace we will have only one possible interpretation.

## 6. Events

We will examine two types of events: relevant and irrelevant, and three groups of relevant events: visible, invisible and semi-visible.

### 6.1 Visible events

These will be the events which depend only on the trace of life (the actions and observations). They depend on what we have already seen and on what we are going to see.

**Note:** Why do we say that an event may depend on the future? Because there will come a moment when the future becomes part of the past and because future can be predicted. Thus, a visible event will become a seen event at some later moment and can be predicted with a certain degree of confidence even before we have seen it. The future is unknown from the perspective of the current moment, but from the perspective of another moment it may already be known. For example, from the perspective of the endpoint of life, all future will already be part of the past and will be known.

Let us have the set *All* of all infinite sequences of actions and observations:

$$\ldots, a_{t-3}, o_{t-2}, a_{t-1}, o_t, a_{t+1}, o_{t+2}, a_{t+3}, \ldots$$

These sequences are infinite in both directions (towards the past and towards the future). Important here is which is the current moment (where does the zero stand). Thus, each infinite sequence can be regarded as a function: $\mathbb{Z} \to \Sigma \cup \Omega$.

**Definition:** Any subset of *All* is a visible event.

In other words, for each infinite sequence of actions and observations and for each particular moment $t$ we can say whether the event occurs or not.

Now, with this definition the visible events have become too many. The members of *All* are continuum many meaning that the visible events are two to the power of continuum.

We do not have the infinite sequence of actions and observations, but only a finite segment (we have the trace of a concrete life). We want visible events to be determined only by this finite segment. Then the visible event will have three possible values: True, False or "Do not know".

**Definition:** A visible event over a segment:

1. Will be True if each and every continuation of the segment to an infinite sequence results in a sequence for which the event is True;

2. Will be False if each and every continuation of the segment to an infinite sequence results in a sequence for which the event is False; and

3. Will be "Do not know" otherwise.

By reducing the visible events to segments we strongly reduced the number of these events. From two to the power of continuum the number is now continuum many. What kind of events were dropped out? Example:

"Event *A* will occur infinite number of times". If we reduce this event to a finite segment, it will always be "Do not know" because we can continue the sequence either way. Certainly, these events are not interesting because we are interested only in those events which we can understand at a given moment.

We will not examine all visible events. Instead, we will examine only the computable ones, withal not all computable ones but only a small part of them. The most important visible events are the atomic ones: *Ob(t)* and *Act(t)*. Composite events which are composed of visible events will also be visible events.

## 6.2 Composite events

We can combine a certain number of atomic events into composite events. The conjunction of atomic events is a composite event.

If we choose to assign a name to the conjunction and convert it into an atomic formula, we will introduce an abbreviating symbol and it will be the name of the conjunction. (We will not observe a disjunction, but can observe the conjunction of negations which is the same thing.) For any composite event we can introduce an abbreviating symbol and start observing it.

We will introduce the conjunction abbreviation symbols by special heuristics which have a confidence coefficient of 100%. The form of these heuristics will be:

$$A_1(t-i_1)\ \&\ A_2(t-i_2)\ \&\ \ldots\ \&\ A_n(t) \Rightarrow Abbreviation(t)\ (100\%),\ i_j \geq 0,\ i_n = 0.$$

We will add *n* more heuristics for the negations:

$$\neg A_j(t-i_j) \Rightarrow \neg Abbreviation(t)\ (100\%)$$

Thus, a conjunction of atomic events will become an atomic event after we determine an abbreviating symbol which designates that conjunction.

Typically, heuristics are obtained through statistics and their confidence coefficient is less than 100%. That is why we will assume that the heuristics introduced here have been added by default.

Another composite event is "*A* will occur before *B*". We will express this event as follows: *exist A after t before B in our case*. This event is visible when *A* and *B* are visible. If in some infinite life events *A* and *B* do not occur after *t*, this composite event will be False.

## 6.3 Invisible events

Here we will discuss Invisible Single-Moment Events (ISME). Certainly, from ISMEs we can obtain other invisible events (e.g. by conjunction).

An ISME is an event which reflects the possible past and the possible future at a particular moment *t*. It will depend on only one state of the world. The interpretation of an ISME will be a set of states of the world.

**Note:** Why is the interpretation of an ISME a set of states of the world rather than a characteristic? Because we consider that the model of the world is fixed and for each state of the world an ISME is either True or False. If we change the model of the world then some of the states of the new world may correspond to a *belief* of states of the previous world. Once we change the world we will also change the interpretation. For each ISME we can choose a model of the world in which the interpretation of this ISME is a set.

The interpretation explains the meaning of an ISME in a concrete world, but we are going to describe ISMEs in any world. Therefore, we will look at the things from the perspective of the agent, not from the perspective of the world.

The possible future can be presented as the tree of all possible developments. We will assume that the possible past can also be presented as a tree. Then the possible past and the possible future can be presented as a tuple of trees. Let *All2* be the set of all such tuples of trees.

**Definition:** Each subset of *All2* will be an ISME.

According to this definition there are too many ISMEs. We will describe six ISME types which are important to us.

### 6.3.1 ALWAYS OCCURS

An example of this type of event is "Event *A* will always occur in future". (Likewise, for the past: "Event *A* has always occurred"). We would like to draw some borderline. That is why we will generalize this event to the following: "In the future, event *A* will occur always until the moment at which event *B* occurs". The concise expression of this event will be *A from t until B.*

It makes sense to try and find some heuristics which predict this ISME. For this purpose we will use statistics over lived experience and will select some conjunction which is a candidate for antecedent of the heuristic. Let this conjunction be True $m$ times and let us assume that in $k$ of these times *A from t until B* has been True. Then $k/m$ appears a good candidate for a confidence coefficient. This candidate however does not reflect the specificity that at a low $m$ values the confidence is also low, so it is better to have $(k-1)/m$. In this case the coefficient will be below 100% but will tend to 100% when $k=m$ and when $m$ tends to infinity.

Let us assume that in our conjunction there is an invisible event. We will also assume that on the basis of certain heuristics this invisible event is True. We can obtain this information from the guesses $G_t$. If there are several guesses for the invisible event, then let $R$ be the highest confidence coefficient possible. In this case we will insert in the counters (in which we accumulate $k$ and $m$) $R$ instead of 1. In this way we acknowledge that the antecedent may not always be True.

### 6.3.2 WILL OCCUR

This type is similar to the event "*A* always occurs", but cannot be obtained from it by two negations because we will be left with the event "*A* may occur". The latter means that *A* will occur in some paths of the possible future, but we want *A* to occur every time (in each possible life).

Accordingly, we will introduce a new invisible event: "*A* occurs before *B*" and will express it in the following way: *exist A after t before B in all cases*.

Again, we can use statistics in order to find some heuristics which can predict this event. We should add here that the predictions should include the case "Do not know". This is the case when life ends before event *A* or *B* has occurred, meaning that we do not know which one (*A* or *B*) would occur if life goes on.

There is some difference between "*in our case*" and "*in all cases*". The first means that the event will occur in our life, while the latter means that the event will occur in each possible continuation of the future. The event "*exist A after t before B in all cases*" is invisible even if *A* and *B* are visible.

### 6.3.3 WILL OCCUR WITH PROBABILITY *P*

The event "*A* occurs between $t$ and *B* with probability $p$" is different from the previous two events. The two events defined above have the properties "retain 0" and "retain 1". When an event "retains 0" this means that when it is False in two states, it is False in any *belief* made of these two states. By analogy, the same applies to the property "retain 1".

The event at hand can be False in two states, but can be True in a *belief* made of these states. For example, let in these two states *A* occur with probabilities $1/4$ and $3/4$. Then from the two states we can make a *belief* in which *A* occurs with probability $1/2$.

### 6.3.4 FEASIBLE EVENT

The event "*A* is feasible" sits between the events "*A* may occur" and "*A* will occur".

"*A* is feasible" means that there is some algorithm *P* such that when the agent executes *P* then *A* will occur. An example of algorithm is $Act(t+1)=a$. It goes without saying that the algorithm can be much more complicated.

How can find some heuristic for "*A* is feasible"? Again, let us select an conjunction which is a suitable candidate for antecedent of the heuristic. Then we will consider different cases. For each case we will find some algorithm *P* and an heuristic which tells us that if the antecedent is in place and if we execute algorithm *P*, then *A* will occur. If we cover all cases we can conclude that if our antecedent is in place then *A* is feasible.

### 6.3.5 TEST EVENT

Test events are discussed in detail in Dobrev (2017a).

An example of a tests event is "If we press the door handle, the door will open". We can call this event "The door is not locked". The value of the test event does not depend on whether we have made the test. Thus, the door may be locked regardless of whether we checked this or not.

### 6.3.6 ED MODEL STATE

Let us have some ED model and try to find the interpretation of that model. For each state $S_i$ of the ED model we will add the event $E_i$ which will be True when the ED model is in state $S_i$. We will try to find a simple interpretation, i.e. we assume that $E_i$ is an ISME.

For each arrow of the ED model we will add one heuristic:

$$E_i(t) \mathbin{\&} A(t) \Rightarrow E_j(t+1) \text{ (per cent)}$$

where *A* is an observed event and the arrow is over that event from $S_i$ to $S_j$.

We will describe the partial trace in a similar way. For each distinction we will add the following heuristic:

$$E_i(t) \Rightarrow A(t) \text{ (per cent)}$$

where $S_i$ is the state, *A* is the distinctive event and the per cent reflects the probability.

With these heuristics we can guess the current state of the ED model and then use the trace in order to predict the additional events.

## 6.4 Semi-visible events

We need to add semi-visible events because of incorrect moves. In order to realize that a move is incorrect we have to try to play it. Accordingly, these events will sit between the visible and the invisible ones. These will be events that we will see if we look.

We will introduce an atomic event (*Correct*) and two heuristics which define that event.

$$Act(t)=a \Rightarrow Correct(a, t) \text{ ( 100\% )}$$

This heuristic tells that if at moment *t* we executed successfully action *a*, then action *a* was correct at that moment. The next heuristic tells us that if at moment *t* we tried unsuccessfully to execute action *a*, then action *a* was incorrect at *t*.

$$UnsuccessfulTry(a, t) \Rightarrow \neg Correct(a, t) \text{ ( 100\% )}$$

The guess *UnsuccessfulTry(a, t)* will be a different guess. Instead of deriving it through heuristics we will add it in $g_t$ by default at each unsuccessful attempt to execute action *a* at *t*. This is the way to add information about unsuccessful attempts.

Why do we add the information about unsuccessful attempts to the guesses? We assume that at any moment the world knows which move is correct and which is incorrect, but does not know which incorrect moves the agent has tried to play. This means that the information about unsuccessful attempts will be visible only by the agent.

The two default heuristics which we added have a confidence coefficient of 100%, but there may be other statistics-derived heuristics which also tell us whether a move is incorrect. It makes sense to assume that when the agent is trained she will come to know which move is correct even without trying to play it. This training will come from the additional heuristics which the agent will derive through statistics.

### 6.5 Irrelevant events

In addition to the relevant events there will be irrelevant ones. These are events which we will ignore and will not bother to understand.

Here are three examples of irrelevant events:

1. “Which life are we living now?” Given that our lived experience is a sequence of many lives, we can ask the question “Which is the serial number of the life we are living now?”

We can assume that the behavior of the world will not change across lives, but it makes sense to assume that in her first life the agent will make certain errors due to lack of experience which she will avoid in the next lives. Nevertheless, we will regard this question as irrelevant and will therefore ignore it.

2. “How old am I and how much time is left for me?” We will ignore this question, too. In other words, we will ignore the parameters *t* and *k-t*. Here is why we ignore this question: We imagine that our agent is eternally young and does not get older as life goes on. As concerns *k-t*, nobody knows how much time is left for them to live.

Later we will define an extended model where *t* and *k-t* matter, but we are only interested in events which if True in one life will remain True in any forward or backward extension of that life. Hence we are only interested in events which do not depend on *t* and *k-t*.

3. We will also regard as irrelevant the events which produce unimportant information. In Section 5.7 we already provided an example in which it does not matter whether the count of occurrences of an event adds up to an even or odd number.

### 6.6 Related work

What is an *event* and who introduced the term *event* in AI? The term was first introduced by Xi-Ren Cao. He introduced *events* in his paper Cao (2005).

Subsequently, in 2008 Xi-Ren Cao and Junyu Zhang provided a definition of the term *event* in their paper Cao and Zhang (2008). That definition is not good enough, because it depends on the model and because it suggests that the last state in which the model has been is remembered, but this is not something worth remembering. This is how their definition of *event* looks like:

$$E \subseteq S \times S \text{ for a special model}$$

$$E = \{ <s_{i-1}, s_i> \mid \text{if } E \text{ happens at moment } i \}$$

Why does this definition depend on the model? Because Cao and Zhang assume that there is only one model of the world, which is not the case. In Dobrev (2019a) we prove that the world has many models and furthermore that there is a Minimal and a Maximal model of the world. *Minimal* and *maximal* in this case refer to what the states of the model know about the past and about the future.

What should the definition be, then?

$$E \subseteq S \text{ for any model}$$

$$E = \{ s_i \mid \text{if } E \text{ happens at moment } i\}$$

In the various models the state can "remember" a larger or a smaller piece of the past. In Xi-Ren Cao's definition the state remembers the last state in which the model has been. As we mentioned already, this is not something worth remembering. If we decide to remember something, it would be better to remember the last action of the agent. This will make it obvious that the agent's actions are events.

While the definition in Cao and Zhang (2008) is not perfect and needs to be improved, this does nowise diminish the credit owed to Xi-Ren Cao because he was the first to realize that observing the actions only is not enough and therefore we should generalize to the wider class of events.

Indeed actions tell us everything, but they provide too much information. Where a model monitors each and every action, it is flooded and overwhelmed by an excessive flow of information. By generalizing actions to events we can reduce the inbound information and stay only with the "important" things.

**Note:** In MDP, observations are not among the monitored events, but they are detected by trace. The trace specifies the current state and thus the observations are taken into account.

**Note:** The term *event* is used in many papers, but usually with another meaning. In Lamperti, Zanella and Zhao (2020) for example, *event* is meant as an *observation*. While an observation is a special case of an event, for us *event* is a more general concept.

## 7. The world

Before creating a language for description of worlds we must be clear about what is a world. We will define the world through its perfect models. The idea is the following: Imagine that you have a painting and a perfect copy of that painting. The copy is so perfect that one cannot tell the copy from the original. In this case we can assume that the original painting and its copy are the same thing.

The world will consist of a model (which will be a Simple MDP) and an initial *belief* which will show us where life is expected to start from. We will want the model to be both perfect (so it cannot be improved any further) and minimal.

While the minimal model is not unique and even the perfect minimal model is not unique, for every pair of two perfect minimal models we can say that the states of the first model can be expressed as *beliefs* of the states of the second model.

### 7.1 The minimal model

A minimal model is one the states of which do not know any redundant facts. That is, they do not remember anything redundant about the past and anything redundant about the future. A redundant (unnecessary) fact of the past is one which does not have any impact on the future. If it does not affect the future, is it worth remembering? A redundant fact about the future is one which does not depend on the past (i.e. cannot be predicted from the past). Does it make sense to have the model know unnecessary (redundant) things about the future? For example, when you receive a letter but have yet opened it, does the world know what does the letter say? We will assume that the past can nowise help us predict the content of the letter. In this case the content of the letter is something redundant for the world, so the world does not need to know it. The world can decide what the content of the letter is only when you open it.

Of course you assume that you live in a real world and the world knows what is in the letter even before you open it. But imagine that you live in a computer program (as in the movie The Matrix) and the world will decide what must happen in the last minute. Thus, the world will decide what is in the letter not at the time it was written but as late as when you open it.

The content of the letter is something redundant for the world, but not for the agent. This fact is important for the agent because it is related to its future, but will be unpredictable by the agent because it is unrelated to the past. Although the fact is unpredictable, the agent will try to predict it because there is nothing to tell her that the content is unpredictable and therefore her attempts will be futile.

## 7.2 The perfect model

As we said, the states of a minimal model do not know anything redundant. Now we say that the states of a perfect model know everything useful. Therefore, the states of a perfect minimal model know exactly what needs to be known (everything useful and nothing redundant).

There are many ways to say that a model is perfect:

1. The model has the Markov property.

2. The future depends only on our state of departure and does not depend on how we have arrived at that state.

3. The model cannot be improved any further into another model capable of making a better prediction of the future on the basis of past.

4. No state can be divided in two states such that the two new states have different past and different future. A different past means that we would be able to distinguish the states on the basis of what happened. To distinguish them means to distinguish with certainty or to say that the one is more probable than the other. It is not mandatory for each possible past to distinguish the two states. Just one possible past that can distinguish the two states is sufficient. By analogy we can say what it means when two states have a different future.

These definitions of a perfect model are all equivalent. The state of the perfect model need not be split in two because it knows everything useful. If we split the state in two on the basis of the past, the two new states will know something more about the past, but that thing will not be useful and therefore the new states will be bound to the same future.

**Question:** If we pick randomly some model, will it be a perfect model of some world? The answer is yes. Every model is a perfect model of some world. Every model describes some world and if we assume that it is the best possible description and that description cannot be improved any further, then the model is perfect. Of course there are infinitely many worlds for which that model provides some partial description which can be improved. The main question is: “Have we found the perfect model exactly of the world we are trying to describe?” The answer to this question is that we do not know. All we know about the world is our lived experience. This experience can occur in infinitely many worlds. Certainly, we are looking for the simplest model that matches our experience (this principle is known as the Occam’s razor).

In fact, we cannot even be sure that the model built on the basis of our lived experience is correct. There are some probabilities in the model which we determine by statistics. When the collected statistics are limited these probabilities may not be accurate, but if we assume that the statistics are sufficient then the only question left would be whether the model is perfect.

We will not try to find a perfect model of the world. All we need is a sufficiently good model which works for us. In Dobrev (2019b) we explained why trying to find a perfect model of the world is an overly ambitious goal. Nevertheless, we will assume that a perfect model of the world does exist and that model provides the definition of the world.

Furthermore, we are not looking for a minimal model of the world because it would make us admit that certain events are unpredictable. As a rule, we will try to predict any event even though we assume that the result from certain events (such casting a dice) is unpredictable.

## 7.3 The basic model

**Definition:** A basic model of the world is every perfect and minimal model of the world.

We will assume that each world has a basic model which provides the definition of that world. A model can be the basic (perfect) one of a certain world, but for another world it may be just an ordinary (imperfect) model.

The states of the basic model know everything useful both about the past (everything that could have happened) and about the future (everything that can happen going forward).

**Note:** Various authors may use different terminology. This, the term used in Schofield and Thielscher (2019) is “game” instead of “world”. Likewise, Schofield and Thielscher use “Game Description Language” instead of “Language for Description of Worlds” as well as “Imperfect Information” instead of “Partial Observability”. We can imagine the world as a game and, vice versa, a game can be thought of as a world in its own right. The Mystery group uses this idea in their song “The World is a Game”.

## 7.4 Simple MDP

Worlds are usually described using Markov decision process (MDP). We will simplify the MDP and create a Simple MDP.

Why does MDP need simplification? Why do we claim that there is a degree of needless complication in the MDP? There are two reasons for this.

First, MDP conceals the fact that besides being a model of the world it also models the agent. Indeed, the world and the agent form a single system, so one cannot describe the first without describing the latter. In other words, MDP describes not only the behavior of the world but also the behavior of the agent. If you look closely at the MDP you will see that according to MDP the agent does whatever she wants. Doing whatever you want is also a kind of behavior, though it is the most liberal behavior possible. What is the probability for the agent to choose a particular action? We do not know – the agent is free to do whatever she likes, so the probability is in the interval *[0, 1]*.

Second, the MDP restrains the world and forces it to follow some fixed policy. I.e., in MDP the agent is unrestrained and free to do anything, while the world is restrained to a fixed policy.

## 7.5 Fixed policy

What is a fixed policy? For example, this is the case when in a given situation you always turn left. Even more so, this is an extreme policy because *always left* and *always right* are two extreme possibilities. But, in a scenario when you toss a coin and if it falls heads then you turn left you will be executing a fixed policy with probability 1/2 (exactly 1/2). If the probability is different then you are executing a different fixed policy. If you turn left or right as you like, then you will not execute a fixed policy and instead will turn left with probability in the interval *[0, 1]*.

Do you have free will (freedom of choice) when executing a fixed policy? The answer is *no*. When you toss a coin, it is the coin which makes the decision, not you. Thus, in MDP the agent enjoys absolute freedom when choosing her move. In MDP the agent is not restrained by anything, while the world is absolutely restrained as it is forced to execute a fixed policy.

## 7.6 Extreme policy

An extreme policy is a fixed policy in which every probability is set to its minimum or to its maximum value.

How can we select an extreme policy? If the probabilities are determined by the intervals $[a_i, b_i]$, we will select the extreme value of one of these intervals (i.e. we select either *a* or *b*). If we select *b*, this may shorten the remaining intervals. Then we will select one of these remaining intervals (probably shortened) and so forth. In this way we select a fixed policy where every probability is extreme (i.e. it cannot be increased any further or it cannot be reduced).

When we make a reference to policy in Simple MDP, we will mean a policy shared by the agent and the world. To put it another way, the two of them have colluded to implement two policies, the combination of which is the policy of the Simple MDP. In the MDP, references to a policy are in fact references only to the policy of the agent because in the MDP the world is bound to a fixed policy and has nowhere else to go. Then, the agent's policy in the MDP is in fact an extreme policy (each one with probability either 0 or 1). What makes extreme policies sufficient in the MDP? When we purse a certain goal and have to decide *left or right*, typically there are three scenarios: 1) our goal will be better served when we go left; 2) our goal will be better served when we go right; and 3) it does not matter whether we go left or right. If we assume that in the third scenario we choose to go left, then we end up with an extreme policy. This makes extreme policies sufficient in the MDP. Certainly, this does not apply to each goal. If the goal is diversity (tour around as many states as possible), then we should alternate left and right.

The good thing about extreme policies is that when the states are finitely many, then the extreme policies are also finitely many, while the fixed policies usually are continuum many.

## 7.7 From MDP to Simple MDP

The MDP can help us describe only part of the worlds. These are some very peculiar worlds where the nature does not have free will as it is forced to execute a fixed policy. If an agent with free will lives in such a world, the world cannot be described by MDP. Let us take the world of the chess game where is a second agent who plays against the protagonist. Let us assume that the second agent is not deterministic (she is not bound to play some extreme policy). Suppose that she even does not play a fixed policy. So, we let the second agent loose and she can play whatever she wants. This type of world cannot be presented by the MDP, but we can describe it using the Simple MDP. The difference between the two models is that instead of exact probabilities, in Simple MDPs we operate with probability intervals.

The MDP is needlessly complicated because it obscures the fact that the probability of the agent's actions is actually the probability interval $[0, 1]$. While the arrows over actions come with probabilities, rather than determining the probability of the action these probabilities determine indirectly the probability of the observation. (The observation has some influence to what the next state of the MDP model will be, but that influence is very subtle and indirect. The arrows define, in a non-deterministic manner, several possible states and some of these possibilities are dropped out on the basis of the observation. With Simple MDP things are much more straightforward because the Simple MDP includes arrows over observations which directly indicate the influence of the observation on what the next state will be.)

In MDP the state has past, present and future. That is, something has happened before the state, something is happening within the state and something will happen after the state. The Simple MDP state has only past and future because nothing happens within the state. In the MDP state there is some observation. There are two equivalent definitions of MDP. We will name them *monochromic* and *polychromic*. With the monochromic definition in the state there is only one possible observation, while the polychromic definition says that in each state there are several possible observations and each observation comes with a fixed probability. This means that the model in the monochromic definition is not minimal because the state "knows" exactly which observation will come up and that knowledge may not follow from the past. It may be the case that the past provides for a non-deterministic transition between several monochromic states. Minimality will occur if these several monochromic states are replaced with a single polychromic state wherein each observation comes with its probability.

## 7.8 Definition of Simple MDP

We will present the Simple MDP as a finite automaton (except for the requirement that the number of state is finite). More precisely, we will present it as a probabilistic automaton because over the arrows we will have probabilities. Even more precisely, we will present it as interval-valued probabilistic automaton because over the arrows instead of probabilities there will be probability intervals.

With the MDP there is only one type of state due to the fact that in MDP it is always the agent's turn to make a move. The world in Simple MDP will be a game between the agent and the world. Thus, there will be two types of states of the world: 1) states in which it is the agent's turn to make a move and 2) states in which it is the world's turn to make a move.

Each arrow in MDP stands for one move. This move comprises some action by the agent and some response by the world. (The response of the world is the observation which the agent sees.) In Simple MDP each arrow will represent a ply (half move). This half move is either the agent's action or the world's response.

**Definition:** Simple MDP is a graph $(U \cup W, A \cup O)$ with two types of nodes and two types of arrows:

- $U$ are the states of the world when it is the agent's turn to make a move (the agent will execute some action);
- $W$ are the states of the world when it is the world's turn to make a move (the world will display some observation to the agent);
- $A$ are arrows which represent actions ($A$-type arrows);
- $O$ are arrows which represent observations ($O$-type arrows);
- If $u \in U$, then the arrows departing from $u$ are $A$-type arrows and the arrows ending in $u$ are $O$-type arrows;
- If $w \in W$, then the arrows departing from $w$ are $O$-type arrows and the arrows ending in $w$ are $A$-type arrows.
- $A \rightarrow \Sigma \times [0, 1] \times [0, 1]$, meaning that each $A$-type arrow is associated with some action and a probability interval;
- $O \rightarrow \Omega \times [0, 1] \times [0, 1]$, meaning that each $O$-type arrow is associated with some observation and a probability interval;

We must add something about the probability intervals. When the probability is fixed (the length of the intervals is zero), then the sum of the probabilities of the arrows which depart from one node must be 1. When the probability is not fixed, then we will consider the probability intervals (of the arrows departing from one node) as a description of a set of vectors of fixed probabilities, wherein the sum of each of these vectors is 1. This means that the intervals $[a_i, b_i]$ describe vector $p_i$, where $a_i \leq p_i \leq b_i$ and $\Sigma(p_i) = 1$. We will want this description to describe at least one probability vector (i.e. the set of the described vectors must not be the empty set). We will also want the description to be optimal (meaning that each shortening of the intervals will remove some vector from the set). In Dobrev (2017b) there are several inequalities which must be true for the description to be both non-empty and optimal.

### 7.9 MDP as Simple MDP

How can we present MDP as a Simple MDP? In MDP all arrows are over actions, which makes them *A*-type arrows. Each arrow will remain in the Simple MDP, the only difference being that its probability *p* will be replaced with the interval *[0, p]*. Hence, if the agent chooses that action, then she will choose that arrow with probability *p* and in the opposite case will choose that arrow with probability 0. Where the action is only one (no arrows over other actions depart from the state), then *p* is replaced with the interval *[p, p]*.

How do we change states? Each state *s* is replaced with two states (*w* and *u*) which belong to type *W* and accordingly to type *U*. All arrows which previous arrived at state *s* will now arrive at state *w* and the arrows which previously departed from *s* will now depart from *u*. We will add additional arrows from *w* to *u*. The number of these additional arrows will be equal to the number of possible observations *ob* in *s*. Each such arrow will be of type *O*, will be over the corresponding observation *ob* and will be associated with the probability interval *[p, p]*, where *p* is the probability of the observation *ob* in *s*. What shall we do in the case of the monochromic definition (where each state has one and only one observation)? Then there will be only one arrow from *w* to *u* and its probability interval will be *[1, 1]*.

The resulting Simple MDP will describe the same world as will the MDP. If the MDP model has been perfect, then the resulting Simple MDP will also be perfect. The same applies to minimality.

So we have proven that all worlds which can be described through MDP can also be described through Simple MDP but not vice versa. In other words, Simple MDP expands the notion of *world*.

### 7.10 Initial *belief*

Before we can describe the world we need to add an initial state. We prefer to have a set of possible initial states instead of a single initial state. Each of these possible states comes with some probability. If this probability is fixed, we will have a structure which we will designate as fixed *belief*.

**Definition:** Fixed *belief*:

- $M \subseteq U \cup W$
- $M \rightarrow [0, 1]$

This means that we have a set of states and have matched each state in that set to a fixed probability. The sum of these fixed probabilities must be 1. In this case saying that $s \notin M$ is not much different from saying that the probability of *s* is zero. In most papers our notion of fixed *belief* is called *belief*, but here *belief* will be something more complex.

**Definition:** A generalized *belief* is the set of fixed *beliefs*.

For the sake of simplicity instead of generalized *belief* we will use just *belief*.

We will assume that the world is described by one Simple MDP and one initial *belief*. Which initial state do we expect to depart from? First, from the members of the initial *belief* we pick one fixed *belief*. How do we pick it? Any way we wish, here we have randomness with unknown probability. Then from the selected fixed *belief* we pick a concrete initial state with the probability given by that fixed *belief*.

The meaning of the notion *initial belief* can be inferred from lived experience and from the statistics of the world. This is the expectation about what the initial state of the world will be.

## 8. Interpretation

We already explained what is an interpretation of an ED model, but have not yet explained the interpretation of an event.

We will introduce the extended model. We will obtain it by adding the past and the future to the basic model. Thus, while the states of the basic model know what could have happened and what can happen going forward, the states of the extended model will know what exactly has happened so far and what exactly will happen from this moment onwards.

The interpretation of an event will be a set of states of the extended model. Each event will have its own probability because each state of the extended model will have its own probability.

In order to simplify the explanation, we will assume that the world is cyclic. The property of being cyclic is characterized by two things:

1. Each event that has already happened may happen again.

2. The initial *belief* is the invariant *belief*. Thus, the question "In which state do we expect to be at the initial moment?" coincides with the question "In which state do we expect to be at a random moment?"

The good thing about the cyclic world is that in it the backward probability does not depend on *t*. Another advantage of the cyclic world is that in it we can define a probability for each state of the basic model and that probability will not depend on *t* either.

### 8.1 Possible future

The possible future is a sequence of observations, actions and sets of incorrect moves. These sets are finite and that is why the possible future is a word over an alphabet composed of $n+m+2^m$ letters.

For each state *s* we can describe the possible future as an infinite tree of all words which the future can produce if we start from *s*. Each node of the tree will be associated with some word (possible future) and some probability (probability interval).

The probability is the product of the probabilities which we have captured in the Simple MDP model when we start from *s* and read the word in question. (Arrows are associated with actions and observations, while *U*-type states are associated with sets of incorrect moves.) If the word in question can be read in two or more ways, then the probability is equal to the sum of the various probabilities which may be obtained by the different readings.

Where the probability is an interval of probabilities we should explain how probability intervals are added and multiplied.

$[a_1, b_1] . [a_2, b_2] = [a_1.a_2, b_1.b_2]$ (Multiplication)

$[a_1, b_1] + [a_2, b_2] = [a_1+a_2, min(b_1+b_2, 1)]$ (Addition)

How the tree of the possible future looks like? The number of arrows over one action which depart from one node cannot be more than one, but the number of arrows over one observation can be more than one. When all moves are correct then the tree is simple and the arrows over observations cannot branch out, either. The presence of incorrect moves makes the tree more complicated because from one node there can depart several arrows over one and the same observation, but they will end in nodes which correspond to different sets of incorrect moves.

We will assume that none of the arrows is associated with a probability of zero. If there is such probability, we will remove that subtree. If we let some arrows have zero probabilities, then the tree will not be unique.

### 8.2 Possible past

We prefer to reuse the definition of the possible future and apply it to the possible past. However, we have a problem here. For each arrow we will need its backward probability, that is the probability of arriving from that arrow. (We already have the forward probability, but it means something else: how probable it is that we departed over that arrow.)

We can define the backward probability by the statistics of the world. Its value will be *k/m*, but *m* here will express the number of times we have been through the head of the arrow (not through its tail). The problem here is that in defining the forward probability we assumed that it is permanent and does not depend on *t* (i.e. does not depend on the step at which we are). However, when defining the backward probability it may turn out that it depends on *t*.

In order to calculate the backward probability, we will need the forward probability and something else, namely the probability of each state. In the initial moment we can present that probability through an initial *belief*. For each next moment we can calculate the next *belief* by using the initial *belief* and the forward probability.

The formulas for this purpose are:

$$m_i . p_i = m \,.\, q_i$$

$$q_i = \frac{m_i . p_i}{m}$$

$$m = \sum m_i . p_i$$

Here $q_i$ are the backward probabilities of some state $s$. The index $i$ runs the previous states (the states of departure of an arrow which arrives at *s*). The probabilities $p_i$ are the forward probabilities of the arrows which depart from the previous states and arrive at state *s*. The probabilities $m_i$ are obtained from the previous *belief*. These say how likely it is that we have been in some previous states at the previous moment, while the probability *m* says how likely it is that at the next moment we will be in state *s*.

The basis of the first equality is that the probability of departing from a certain state over a certain arrow is equal to the probability of arriving at the next state over the same arrow. The formulas above determine the backward probabilities $q_i$ in all cases except one. This is the case when $m=0$. In this case we can assign to $q_i$ any values we wish (as long as the sum of $q_i$ is 1).

Thus, for each step *t* we obtain some *belief* and backward probabilities which depend on step *t*. We defined the possible future as an infinite tree, but the possible past will be much more complex. It will be a countable set of trees. At moment *0* we will not have any past (i.e. a tree with a depth of 0). The possible past at moment *t* will be presented as a tree with a depth of *t*.

Preferably, the possible past should be as simple as the possible future. For this purpose the expectation about the state in which we are must be constant (independent from step *t*).

### 8.3 Past vs Future

In this paper the past and the future are absolutely symmetric. This symmetry comes from the fact that we present life as a path in a certain graph. The direction we follow does not make a difference, i.e. moving forward in the direction of the arrows is not any different from moving backward against the arrows.

At a first glance it appears that there is some difference between forward and backward probabilities, but in fact there is not any difference between these probabilities, either. The forward probabilities are constant, while the backward probabilities depend on *t*, but this is because we first establish the forward probabilities and thereafter establish the backward probabilities. With the inverse approach (establish the backward probabilities first and on their basis establish the forward probabilities) the backward probabilities would be constant and the forward probabilities would depend on $(k - t)$. Certainly, when we walk backwards we need to have a final *belief* instead of an initial *belief*. Hence, we must depart from some final *belief*.

Later on we will introduce the notion of *cyclic world*. This will be a special case where the forward and backward probabilities do not depend on $t$. In this way we will demonstrate that in the interesting case the past and the future are absolutely symmetrical.

### 8.4 Free will

We assumed that the world and the agent may have free will (freedom of choice). I.e. we suggest that the world and the agent can do whatever they like (but within certain limits). Since we describe the world through statistics, the players must make use of their freedom, otherwise the model will not be able to capture that freedom.

If the agent and the world do not exercise their freedom to the fullest extent, these statistics will produce some abridged model of the world which reflects only the actual behavior of the players.

For example, in a chess game the agent and the world have some correct moves and can play any of them. But in an actual game they will not play each and every correct move, e.g. we may reasonably suggest that they will not make stupid errors. If we use statistics to describe the chess game, the probability intervals perhaps will be shorter, i.e. the resulting model will not capture all of the possible moves. This will reflect the actual behavior of the agent and of her opponent (there are possible moves which the two of them will never play).

In chess there is one rule which says: "You cannot make a move if it enables the opponent to capture your king in the next move". We can say that such a move would be a stupid error and the rules of the game prohibit stupid errors. So, this rule reduces the number of correct moves. If we prohibit other stupid errors we will end up with a more abridged model in which the behavior of the world and agent will be even more restricted.

### 8.5 Invariant *belief*

We wish to have an initial *belief* such that the next *belief* is the same and every next *belief* is the same. I.e. we wish the initial *belief* to be an invariant one (an invariant point).

We will assume that the probability interval of each arrow is different from 0 (i.e. the interval can be $[0, p]$ but cannot be $[0, 0]$).

**Definition:** A trap is a group of states which one can enter but can never exit.

If there are not any traps in the world, then each connected component will be a strongly connected graph. Let us assume that in each connected component we have selected one strongly connected component to be the main one (let's call it the kernel of the connected component). There are not any traps in the world if and only if there are not any inflows and outflows.

**Definition:** Inflow is a group of states from which there is a way into the kernel, but no way back.

**Definition:** Outflow is a group of states to which there is a way from the kernel, but from which there is no way back.

For an invariant *belief* to exist, the probability of coming from an inflow has to be zero. Otherwise, the probability of being in the inflow will decrease, but it has to be constant. Therefore, for invariant *belief* to exist, there must be no inflows. Analogically for outflows.

## 8.6 Cyclic world

Let us first define the concept of compact world:

**Definition:** A compact world is a world without any inflows and outflows.

**Definition:** A cyclic world is a world which satisfies the following three conditions:

1. There is a path between any pair of states (strongly connected graph);

2. The probability interval of each arrow is different from 0; and

3. The initial *belief* of the world is the set of all invariant fixed *beliefs*.

The first condition means that the cyclic world is compact and has one connected component. The second condition ensures that we can go into every state with some probability if there is some support from the world and the agent (through their free will). The third condition means that the following two questions have the same answer:

1. Which state do we expect to depart from?
2. In which state do we expect to be after some very long life, regardless of our state of departure?

We will assume that the world is compact and even cyclic. The good thing about the compact world is that the backward probabilities are unambiguously defined so that the possible past is simple (it is one tree). The good thing about the cyclic world is that we can define unambiguously the invariant *belief* and thereby provide a more simple definition of the extended model. In the opposite case the expectation as to which particular state we are in will depend on connected component and on the way in which the initial *belief* distributed the probability over different connected components.

In Dobrev (2000) we assumed that our objective is to find some AI which will work well in worlds that are free from fatal errors. The assumption in Dobrev (2000) was that if some AI works well in this kind of worlds, it would work well in any world. The same assumption can be made in respect of the cyclic worlds.

Can we say that a cyclic world is free from fatal errors? In the next sections we will see that if a world is cyclic and is not hostile, then any error would not be fatal. What do we mean by a hostile world? We said that the world enjoys some freedom of choice so it can choose its strictly defined policy from a pool of strictly defined policies. A hostile world can intentionally choose a policy which works against the agent. Then the agent may end up in some concavity without being able to crawl out of it because the world would not let her get out. If the world is not hostile (i.e. if it is impartial or even good-minded and helpful), then the agent would be able to get of any concavity it falls in so there will not be any fatal errors.

## 8.7 A unique *belief*

When is the invariant fixed *belief* unique? Let's look at the more simple case when we have a fixed policy (i.e. the probabilities $p_i$ are fixed). Let also the probabilities $p_i$ be different from zero. And finally, let the world be cyclic.

In this case we would have a unique invariant *belief* (that belief will be fixed). We can calculate it by solving a system of equations. Another way to obtain this fixed *belief* is through the statistics of the world. We can pick some very long life. It does not matter which is the initial state. Let then *m* be the probability of us being in state *s*. That probability will tend to the value given to us by the unique *belief*. Here we do not mean the probability at moment *t*, but the average probability. The reason is that the probability may have a waveform pattern and at the moment *t* it may be divergent, while the average probability is convergent.

So we obtained an invariant *belief* which answers the natural question: “Which state do you believe the world is in?”

Let us now the world be compact, but not cyclic and let there be two connected components. Then the invariant *belief* will not be unique because we can distribute the probability of the two connected components in any ratio (e.g. *1:1* or *1:2*).

Now let the world be cyclic, but with a fixed policy where some probabilities are equal to zero. Then this policy will divide the world in concavities with flat bottoms and surroundings around the concavities. Once we hit the bottom of a concavity we will never be able to get out of it. If we depart from some surrounding around a concavity will end up in the bottom of that or of some other concavity. In this case the invariant *belief* will not be unique because the probability in the surroundings around the concavities will be zero (it will be distributed across the bottoms of the concavities). The problem is that the probability can be distributed in any ratios across the bottoms.

How can we in this case construct a unique *belief*? Let us look at the limit where the probabilities $p_i$ are different from zero, but tend to the ones we have. In this way we will find the natural distribution of the probabilities over the various bottoms. This is how we can find a unique invariant *belief* when the policy is fixed.

If we take the set of all fixed policies (which fall within the probability intervals) we will obtain a set of fixed *beliefs*. That will be the generalized *belief* which answers the question: “Which state do I believe I am?”

For each state of basic model the generalized *belief* will return one probability interval. That interval will tell us how likely it is that we are in a certain state of the basic mode (the probability of us being in that state). Let’s designate this interval with the function *expected*.

### 8.8 Extended model

We said that the states of the basic model know everything useful and nothing redundant. In the extended model we will add one piece of redundant information. We will add the entire life (the trace and the backbone). This information is “redundant” because the state of the basic model “knows” what is the possible past and the possible future, but does not know what exactly happened in the past and what exactly will happen in the future.

The extended model will have a set of states $S'$.

$$S' = \{ < t, L> \mid 0 \leq t \leq \mathrm{k}, L \textit{ is possible life}\}$$

Here $k=|L|$. What do we mean by “possible life”? This is any finite path in the graph which describes the basic model. (Any state can be the initial state because of our assumption that the world is cyclic.)

The set $S'$ will be the *set of moments* because each moment in each possible life is a member of that set.

If we present the extended model as a graph it will have a very simple structure composed of non-intersecting threads. Each thread will be associated with one possible life.

The state of the extended model will know what exactly has happened and what exactly is going to happen, however, this does not mean that if we know the past and the model we will be able to foretell the future. While it is true that the state “knows” everything, we cannot know which particular state we are in.

When $t=0$, then the initial *belief* define the particular state of the extended model we are in. We will obtain the new initial *belief* from the set $S_0'$ by adding to each state the probability *expected*($<0, L>$) (these definitions are provided below).

$$S_0' = \{ < 0, L > \mid L \textit{ is possible life} \}$$

Thus, the new initial *belief* will contain the initial moment of each possible life. These are many members because even if fix the initial state $u_0$, then the members of the new initial *belief* will again be countably many, because the possible lives starting from $u_0$ are countably many (due to the variety of lengths and branches). This means that state $<0, L>$ "knows" everything about the future, but we do not know it because we have no way to know that we are exactly in state $<0, L>$.

We will now extend the function *expected*, which is defined for the states of the basic model, by defining it for the moments (i.e. the states of the extended model). For this purpose will take the probability of life $L$ and will divide by the number of moments in $L$.

We will want the sum of the probabilities of all lives to be 1 and that is why we will assume that the life can be of length $k$ with a probability of $(1-\lambda).\lambda^k$. (So we distribute 1 across all possible lengths of the life.) We will select some coefficient $\lambda$, however, the way we choose that coefficient is not important because we assumed that events are truth-preserving under life extension. (The formula for *expected* will depend on $L$ but not on $t$ because the probability is distributed evenly across all moments of life $L$.)

$$expected(< t, L >) = \frac{expected(s_0).p_1.p_2. \dots .p_k}{k+1}.(1-\lambda).\lambda^k$$

Here $k=|L|$, $s_0$ is the first state of life $L$ and $p_i$ is the probability of the arrow from $s_{i-1}$ to $s_i$. Let $q_i$ be the backward probability of the arrow from $s_{i-1}$ to $s_i$. Then we have three ways to express the probability that some life of length $k$ is exactly life $L$:

$$expected(s_0).p_1.p_2. \dots .p_k$$

$$q_1.q_2. \dots .q_k .expected(s_k)$$

$$q_1. \dots .q_t .expected(s_t).p_{t+1}. \dots .p_k$$

That is, we can start from the probability of any state along the backbone of life and multiply it by the probability of transitioning to the next (or the previous) state.

**Definition:** The interpretation of event *A* is the tuple of states *P* and *Q*, where *P* are the moments at which *A* occurs, *Q* are the moments at which *A* does not occur, and the remaining moments are those for which we do not know whether A occurs or not.

**Definition:** The probability of event *A* is:

$$expected(A) = \frac{expected(P)}{expected(P) + expected(Q)}$$

This means that the probability is determined by the moments at which we know what the value of the event is.

We assumed that events are truth-preserving under life extension, which translates in the assumption that the sets *P* and *Q* are closed under the "life extension" operation.

Our definition of event depends on the selected basic model. A different basic model would come with different $<P, Q>$ tuple. Nevertheless, the event will be the same because it will have the same probability and will behave in the same way versus the trace of life.

### 8.9 MDP as an ED model

As an application of the above we will prove that MDP is a special case of an ED model. With MDP models the observed events are the agent's actions:

$$act(t)=a_i, 1 \leq i \leq m$$

These events do not intersect so we do not need any rules for the reconciliation of collisions. The additional events (those which participate in the partial trace) are the current observations:

$$ob(t)=o_i \vee ob(t+1)=o_i,\ 1\leq i \leq n$$

Here it is not clear whether $t$ is an even or odd number and whether the current observation has just occurred or will occur at the next step. For this reason the event is defined by disjunction.

### 8.9.1 FULLY OBSERVABLE MARKOV DECISION PROCESS

Let's first look at the Fully observable MDP (FOMDP). In this case the topology of the ED model is a simple one. The model has $n$ states (the number of possible observations) and there is an arrow over each action between each pair of states. The trace of the ED model is clear and nice. In each state there is one current observation which occurs with a probability of 1 (each state is identified unambiguously by the current observation). The only thing we need to do in order to define the FOMDP model are the probabilities over the arrows:

$$p_{iaj} = expected(ob(t+2)=o_j \mid ob(t)=o_i \ \&\ act(t+1)=a\ ) \quad (1)$$

This formula says that the next observation is $o_j$, given that the previous observation has been $o_i$ and the last action has been $a$. Here we should explain what does it mean that an event is under a condition. This means that the event will be True or False when the condition is fulfilled and "Do not know" when the condition is not fulfilled.

The function *expected* can return either some strictly specified probability or some probability interval. When *expected* returns strictly defined probabilities in (1), we will obtain a classic FOMDP definition. Certainly, nothing prevents us from generalizing the FOMDP definition and thus allow for transitions the probability of which will be probability intervals.

**Note:** The FOMDP has a single unambiguous interpretation and it is defined by the event "What is the current observation?" We know what is the current observation for any moment of life except for the first moment (moment zero). Not knowing the current observation at the first moment is not a problem because the interpretation need not be defined for a finite life. It must be defined over any infinite life, and infinite lives do not have a first moment.

Thus we have obtained a model with $n$ states and a transition probability matrix $p_{iaj}$. In literature this model is referred to as Full Observable MDP. The question which comes with it is whether the model satisfies the Markov property or in other words is the model perfect (the best possible model which cannot be improved any further).

On one hand, we should be happy if the model satisfies the Markov property because this tells us that we have found the best solution. On the other hand, we should not be so happy because a perfect model means that we have hit a stonewall beyond which it is not possible to improve the model any further.

Accordingly, for each model we will assume that while there is chance for the model to satisfy the Markov property, it is more likely that the model does not. In other words, we will imagine that it might be the best possible model for this world, but in all probability there are better models out there.

### 8.9.2 PARTIALLY OBSERVABLE MARKOV DECISION PROCESS

This is the next model we will deal with. Again, its topology is simple. We have a random number of states and arrows over each action between each pair of states. We have probabilities over the arrows, but also probabilities over the trace. For each state of the ED model we have $n$ distinctions. In each state, each observation will be observed with some probability and, in the general case, that probability will be different from the average one.

The question is whether there is any interpretation of the so-described Partially observable MDP (POMDP). There may or may not be an interpretation. (If the model is inadequate, then there will not be any interpretation.) Therefore we will examine the set of possible interpretations and for each interpretation we will determine the probabilities over the arrows and over the trace.

We will examine the unambiguous interpretations. Thus, we will partition the set $S'$ into equivalence classes ($S'$ is the set of moments). After partitioning $S'$ each state of the ED model will correspond to one equivalence class.

**Note:** The equivalence relation $R$ is not fully random because $R$ must retain the transition over observations. This means that for each odd-numbered moment $<t, L>$ that moment and the next moment $<t+1, L>$ must be in the same equivalence class. A sufficient condition for retaining the transition over observations is to define $R$ on the basis of some equivalence relation over the set $O$ (the set of arrows over observations).

Let events $C_i$ are those which determine the equivalence classes of $R$.

$$p_{iaj} = expected(C_j(t+2) \mid C_i(t) \ \&\ act(t+1)=a\ )$$

Thus, the next state is $j$ if the previous state was $i$ and the last action has been $a$.

What will be the POMDP trace? For each state $i$ we have $n$ possible observations and each one of these observations occurs with some probability $q_{ij}$.

$$q_{ij} = expected(ob(t)=o_j \mid C_i(t)\ )$$

As we said already, the function *expected* can return either some strictly specified probability or some probability interval. If we aim to obtain a classic POMDP definition we have to assume that $p_{iaj}$ and $q_{ij}$ are strictly specified probabilities.

**Note:** If we stay with the standard definition of Partially observable MDP, then the only possible trace of the model will be the probabilities $q_{ij}$. However, we will assume that there may be other traces, i.e. other distinctions, which are characteristic of a given state $C_i$. For example, it may be the case that a certain action can never occur in that state. Another possible distinction is that a certain conjunction of events is impossible, even though the individual events in the conjunction are perfectly possible.

**Note:** The following question arises: "Is there a POMDP which is a perfect model of the world (a POMDP which satisfies the Markov property)?" We have a plethora of POMDPs (for example there is one POMDP for each equivalence relation over $O$). But, among all these models, is there a perfect model? The answer is "Yes, there is." This is how we will obtain that perfect model from the basic model: We will take all arrows in $O$ and for each arrow we will create the event "Walk over this arrow". This event will be True at each odd-number moment when we depart over the arrow as well as at the next moment. The so-obtained POMDP model will be perfect as a sequel the perfection of the basic moment. The model will not be minimal, but if we put together all arrows which have the same beginning and the same end, then we will obtain a perfect and minimal POMDP (as a sequel of the perfection and minimality of the basic model).

## 8.10 Related work

The idea which drives the creation of ED models is that the states of the world a far too many so we should reduce them in order to make the model more understandable. The same idea is embodied in Wang, Joshi and Khardon (2008). Their paper introduces the concept Relational MDPs, which essentially are a special case of the ED models.

We said that any equivalence relation (which partitions the set of arrows) can be used for creating a POMDP. Wang, Joshi and Khardon propose that certain events be selected for a Relational MDP. (Instead of events they refer to predicates, actions and rewards.) These events are then used to define the equivalence relation (two states are equivalent if the same events from the pool of selected events occur in these two states). A POMPD is defined over the quotient set of that relation and this is the wanted Relational MDP.

**Note:** The definition of Wang et al. (2008) does not use a quotient set, but refers to a family of MDPs. If each of these MDPs has only one state from each equivalence class, then it will have the structure of the quotient set, but typically this is not the case. Therefore, it is better to define Relational MDPs through the quotient set.

How does a Relational MDP look like? We will explore three scenarios:

1. Let the input consist of 10 bits. Let these 10 bits be the ten events which have been selected and used for the creation of the Relation MDP. Then the resulting Relational MDP coincided with the FOMDP.
2. Now let only 5 of these 10 bits be selected events. This will give us a Relational MDP with less states than the FOMDP states.
3. Let us now have the selected 10 input bits (events) plus 5 other events which are invisible (do not follow from the input). Let's take these 15 events and create a Relational MDP. Now the Relational MDP will have more states than the FOMDP.

## 9. Description of the chess game

### 9.1 Computer emulation

We have emulated the chess-game world by the computer program (Dobrev, 2020a) written in the language Prolog (Dobrev, 2020b). The rules of the game used by that program are presented as ED models.

When you start the prorgam (Dobrev, 2020a), in the bottom of the screen you will see the visuals provided in Figure 4.

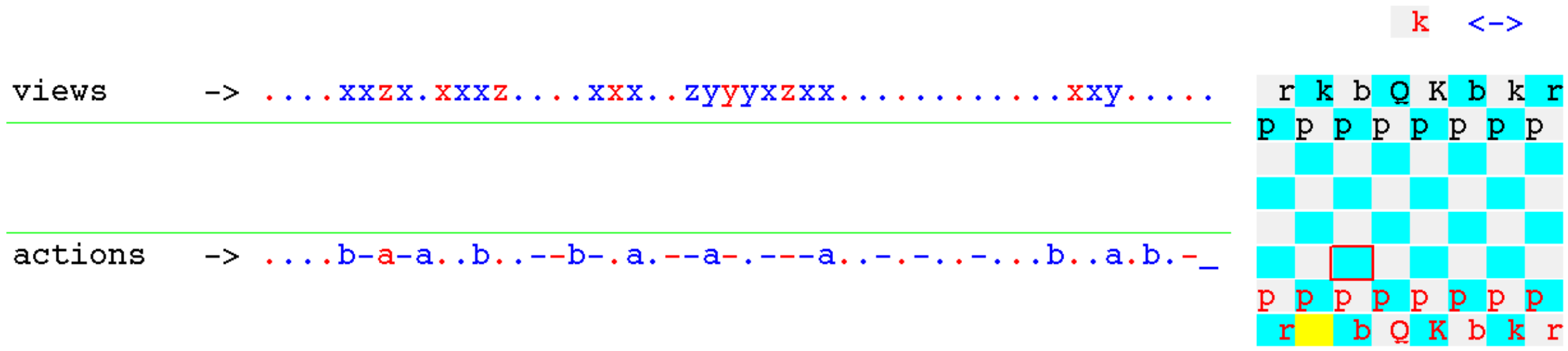

Figure 4

The left-hand side of Figure 4 shows the stream of input-output information, in fact not the full stream, but only the last 50 steps. The top row shows the agent's observations and the bottom row – the agent's actions. There are four possible observations: {0, x, y, z}. The possible actions are also four: {0, a, b, c}. For the sake of legibility the nil and the 'c' character are replaced by dots and minuses.

All the agent can see is the left-hand side of Figure 4. The agent cannot see what is in the right-hand side, and must figure it out in order to understand the world. In the right-hand side we can see (i) the position on the board, (ii) the piece lifted by the agent (knight), (iii) the place from which the knight was lifted (the yellow square) and the square observed currently (the one framed in red).

### 9.2 We use coding

The agent will be able to do 8 things: move hers gaze (the square currently observed) in the four directions, lift the piece she sees at the moment and drop the lifted piece in the square she sees at this moment. The seventh and eighth thing the agent can do is "do nothing".

We will limit the agent's actions to the four characters {0, a, b, c}. The 0 and 'c' symbols will be reserved for the "do nothing" action. This leaves us with 6 actions to describe with as little as 2 characters. How can we do that? We will do that by coding: Let us divide the process in three steps. Every first step will describe how we move the square in horizontal direction (i.e. how we move the observation gauge). Every second step will describe how we move the square in vertical direction and every third step will indicate whether we lift a piece or drop the lifted piece.

We mentioned in Dobrev (2013) that we should avoid excessive coding because the world is complicated enough and we do not want to complicate it further. However, the coding here is not excessive because it replaces eight actions with four and therefore simplifies the world rather than complicate it.

### 9.3 Two void actions

Why do we introduce two actions that mean "I do nothing"? Actually, when the agent just stays and does nothing, she observes the world. The question is, will she be a passive observer or she will observe actively?

When you just stay and observe the world, you are not a passive observer. At the very least, you are moving your gaze.

All patterns that the passive observer can see are periodic. In a sense, the periodic patterns are few and not very interesting. Much more interesting are the patterns that the active observer can see.

We expect the agent to be able to notice certain patterns (properties). For example, the type and color of pieces are such properties. When the agent stays in a square and does nothing, it will be difficult for her to detect the pattern (property), especially since she may have to detect two or three patterns at the same time. If the agent is active and can alternate two actions, then the patterns she observes will be much clearer and more quickly detectable.

To distinguish between the two "I do nothing" actions, we called the second one "surveillance".

### 9.4 One, two, three

The first pattern which will exist in this particular world (the game of chess) stems from our division of the steps in three groups. Let us name this pattern "One, two, three". The pattern is modeled in Figure 5.

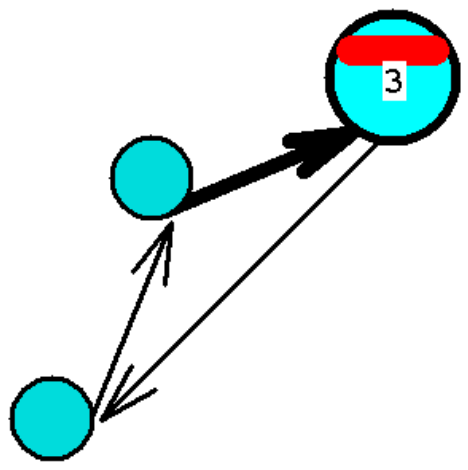

Figure 5

What is the gist of this pattern? It counts: one, two, three.

The pattern is presented through an Event-Driven (ED) model. This particular ED model has three states. The event in this model is only one and this is the event "always" (i.e. "true" or "at each step").

### 9.5 Trace

Does anything specific occur in the states of the above model so that we can notice it and thereby discover the model? In other words, is there a "trace"? This terminology was introduced in Dobrev (2018).

Yes, in the third state, action ***a*** or action ***b*** (or both) must be incorrect. The reason is that the third state indicates whether we lift the piece we see or drop a piece which is already lifted. These two actions cannot be possible concurrently.

We can describe the world without this trace, but without it the “One, two, three” pattern would be far more difficult to discover. That is why it is helpful to have some trace in this model.

The trace is the telltale characteristic which makes the model meaningful. Example: cold beer in the refrigerator. Cold beer is what makes the fridge a more special cupboard. If there was cold beer in all cupboards, the refrigerator would not be any special and it would not matter which cupboard we are going open.

The trace enables us predict what is going to happen. When we open the fridge, we expect to find cold beer inside. Furthermore, the trace helps us recognize which state we are in now and thereby reduce non-determinacy. Let us open a white cupboard, without knowing whether it is the fridge or another white cupboard. If we find cold beer inside, then we will know that we have opened the fridge and thus we will reduce non-determinacy.

We will consider two types of traces – permanent and moving. The permanent trace will be the special features (phenomena) which occur every time while the moving trace will represent features which occur from time to time (transiently).

An example in this respect is a house which we describe as an Event-Driven model. The rooms will be the states of that model. A permanent feature of those rooms will be number of doors. Transient phenomena which appear and then disappear are “sunlit” and “warm”. I.e. the permanent trace can tell us which room is actually a hallway between rooms and the moving trace will indicate which room is warm at the moment.

Rooms can be linked to various objects. These objects have properties (the phenomena we see when we observe the relevant object). Objects can also be permanent or moving and accordingly their properties will be relatively permanent or transient phenomena (a relatively permanent phenomenon is one which always occurs in a given state). Furniture items (in particular heavyweight ones) are examples of permanent objects. People and pets are examples of moving objects. To sum up, a fixed trace will describe what is permanent and a moving trace will describe what is transient.

## 9.6 Horizontal and Vertical

The next Event-Driven model we need for our description of the world is the Horizontal model (Figure 6).

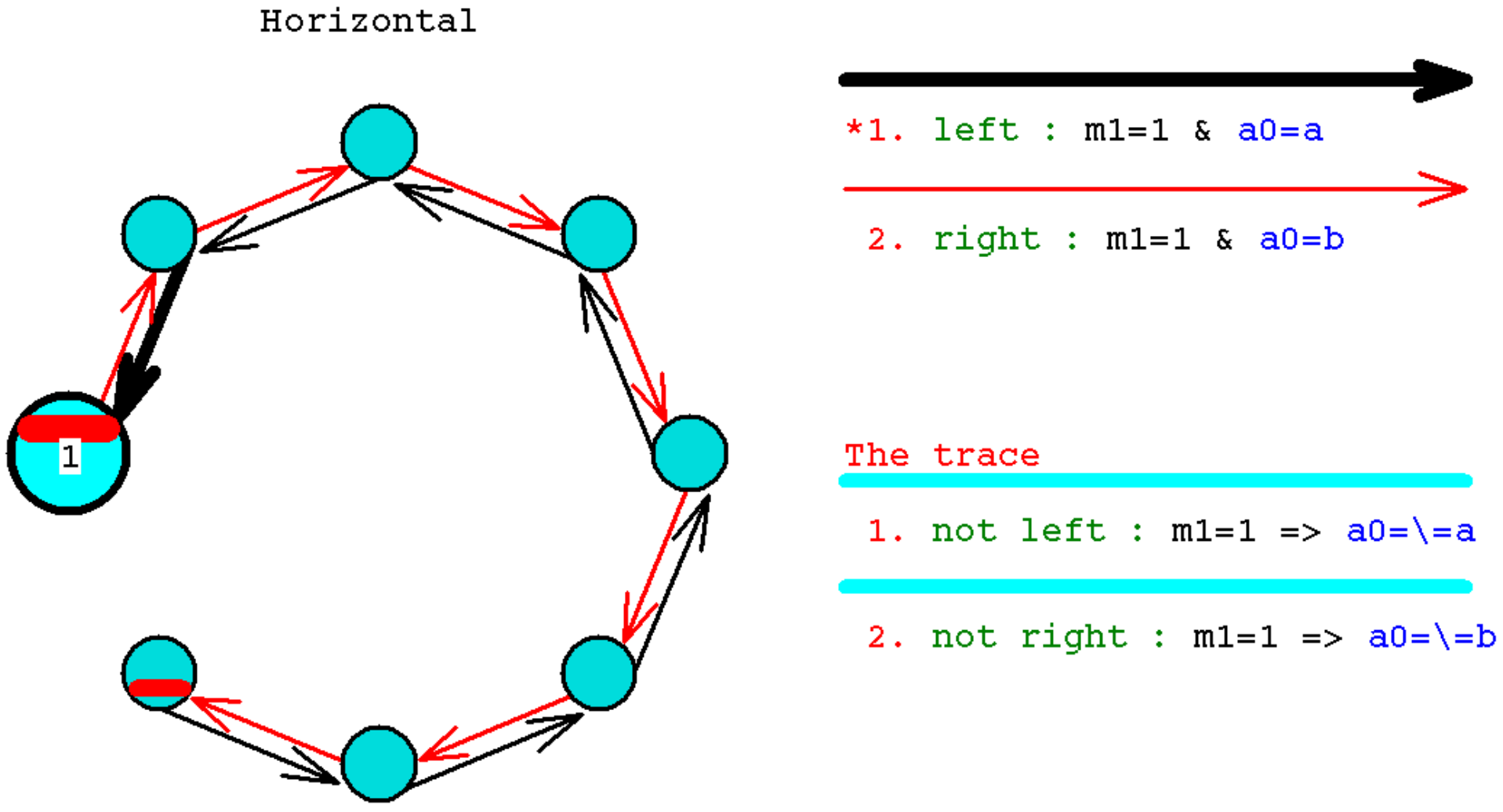

Figure 6

This model tells us in which column of the chessboard is the currently observed square.

Here we have two events: *left* and *right* which reflect the direction in which the agent moves hers gaze – to the left or to the right. So, the agent performs the actions *a* and *b* when model 1 is in state 1. We also have two traces. In state 1 playing to the left is not possible. Therefore, the *left* event cannot occur in state 1. Similarly, we have state 8 and the trace that playing to the *right* is not possible. These two traces will make the model discoverable. For example, if you are in a dark room which is 8 paces wide, you will find that after making 7 paces you cannot continue in the same direction. You will realize this because you will bump against the wall. Therefore, the trace in this case will be the bump against the wall. These bumps will occur only in the first and in the last position.

In addition to helping us discover the model, the trace will do a nice job explaining the world. How else would you explain that in the leftmost column one cannot play *left*.

The next model is shown on Figure 7. This model is very similar to the previous one.

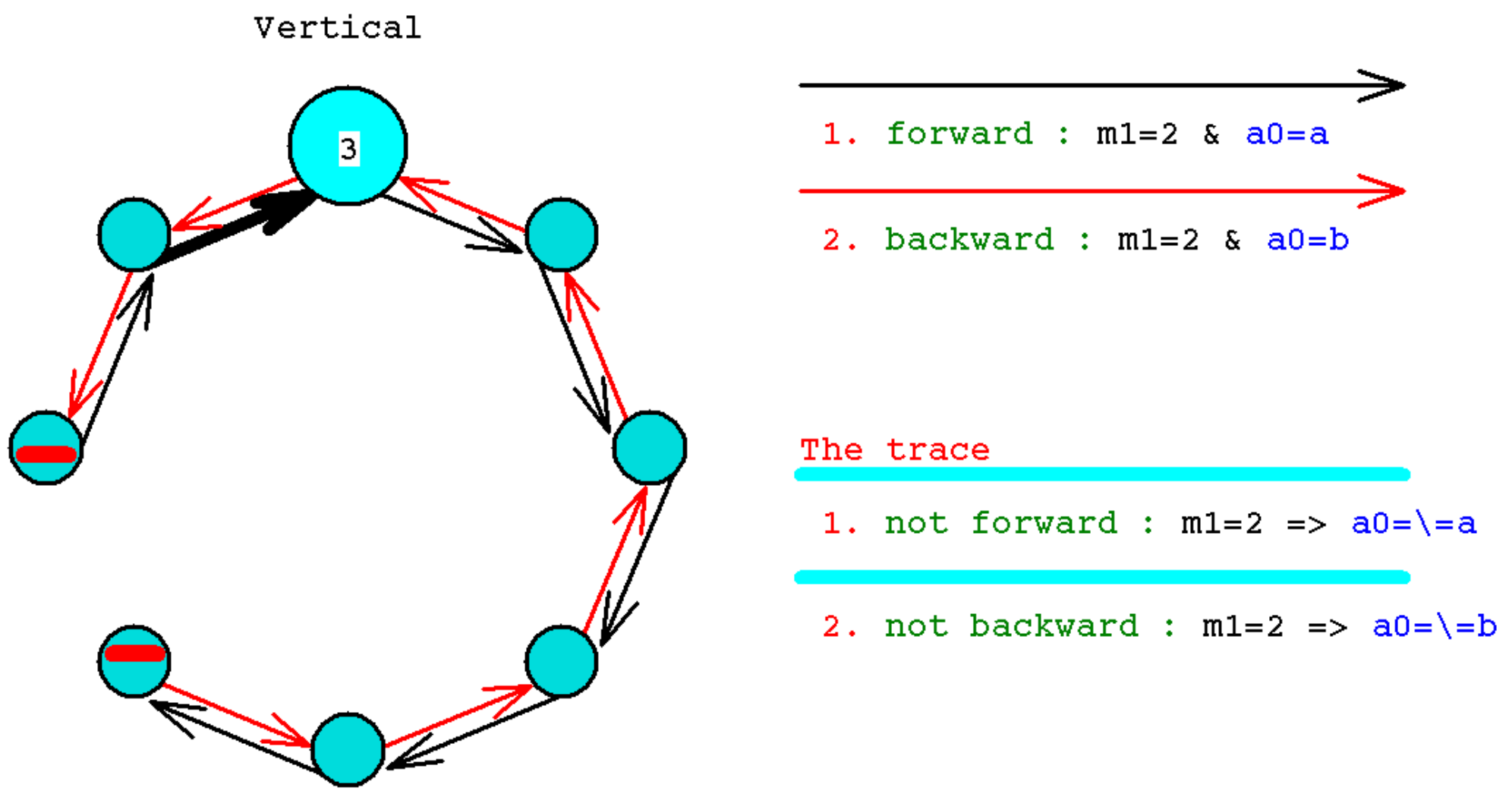

Figure 7

This model will tell us the row of the currently observed square. Likewise, we have two events (*forward* and *backward*) and two traces (*forward move not allowed* and *backward move not allowed*).

It makes perfect sense to do the Cartesian product of the two models above and obtain a model with 64 states which represents the chessboard.

The bad news is that our Cartesian product will not have a permanent trace. In other words, nothing special will happen in any of the squares. Indeed, various things happen, but they are all transient, not permanent. For example, seeing a white pawn in the square may be relatively permanent, but not fully permanent, because the player can move the pawn at some point of time.

Thus we arrive at the conclusion that the trace may not always be permanent.

## 9.7 The moving trace

As we said, moving traces are the special features which occur in a given state only from time to time (transiently), but not permanently.

How can we depict a moving trace? In the case of permanent traces, we indicated on the state whether an event occurs always in that state (by using red color and accordingly blue color for events which never occur in that state).

We will depict the moving trace by an array with as many cells as are the states in the model under consideration. In each cell we will write the moving traces which are in the corresponding state in the current moment. That is, the moving trace array will be changing its values.

Here is the moving trace array of the Cartesian product of models 2 and 3:

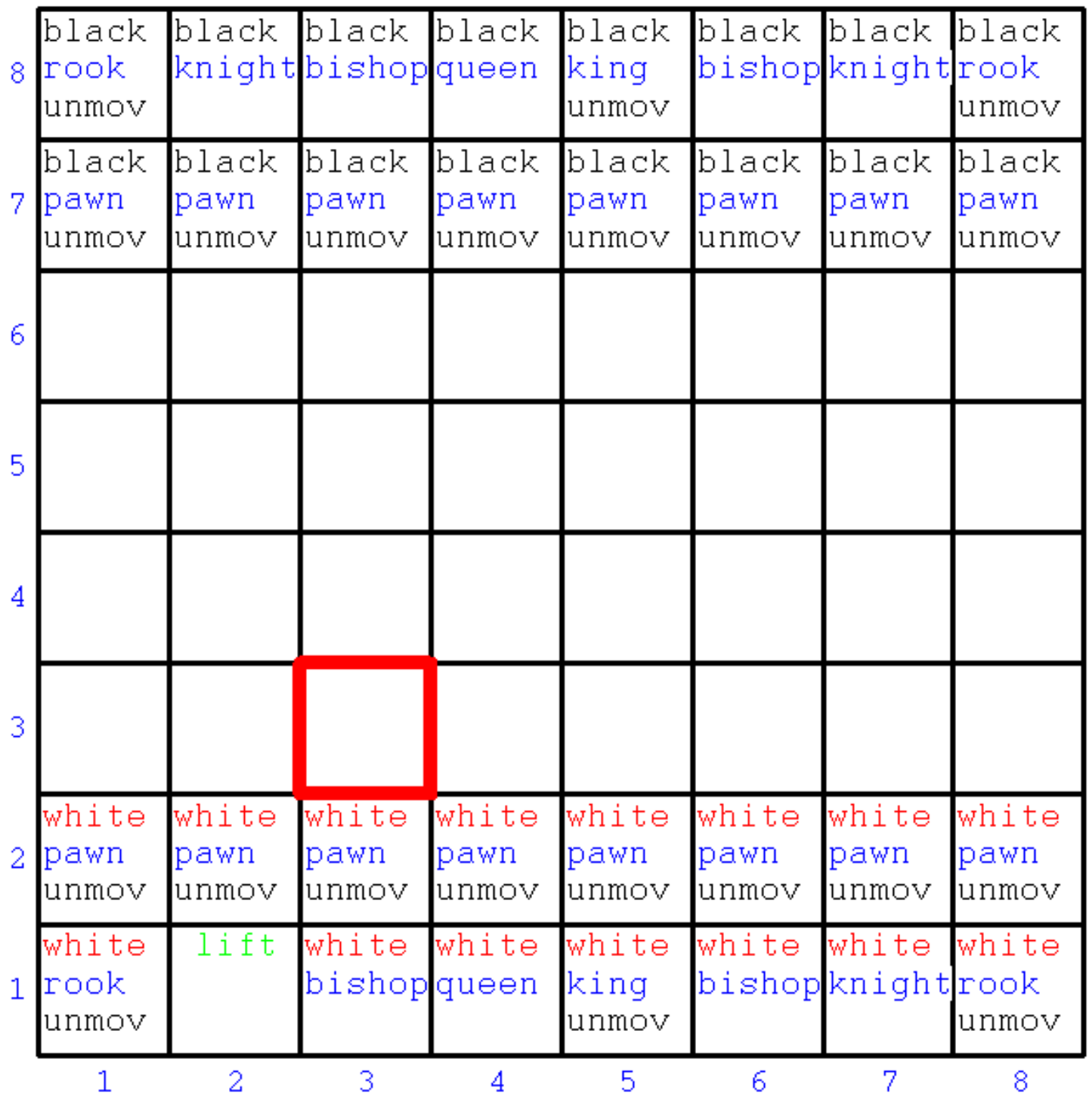

Figure 8

This moving trace is very complicated because it pertains to a model with 64 states. Let us take the moving trace of a model with two states (Figure 9). This is model 4 which remembers whether we have lifted a chess piece. Its moving trace will remember which the lifted piece is. Certainly, the model will also have its permanent trace which says that in state 2 *lifting piece is impossible* and in state 1 *dropping piece is impossible*.

The moving trace of that model will be an array with two cells which correspond to the two states of the ED model. The cell which corresponds to the current state is framed in red. The content of the current cell is not very important. What is important is the content of the other cells because they tell us what will happen if one of these other cells becomes the current cell. In this case, if we drop a lifted piece we will go to state 1, where we will see the lifted piece. (We will see what we have dropped, in this case a white knight.)

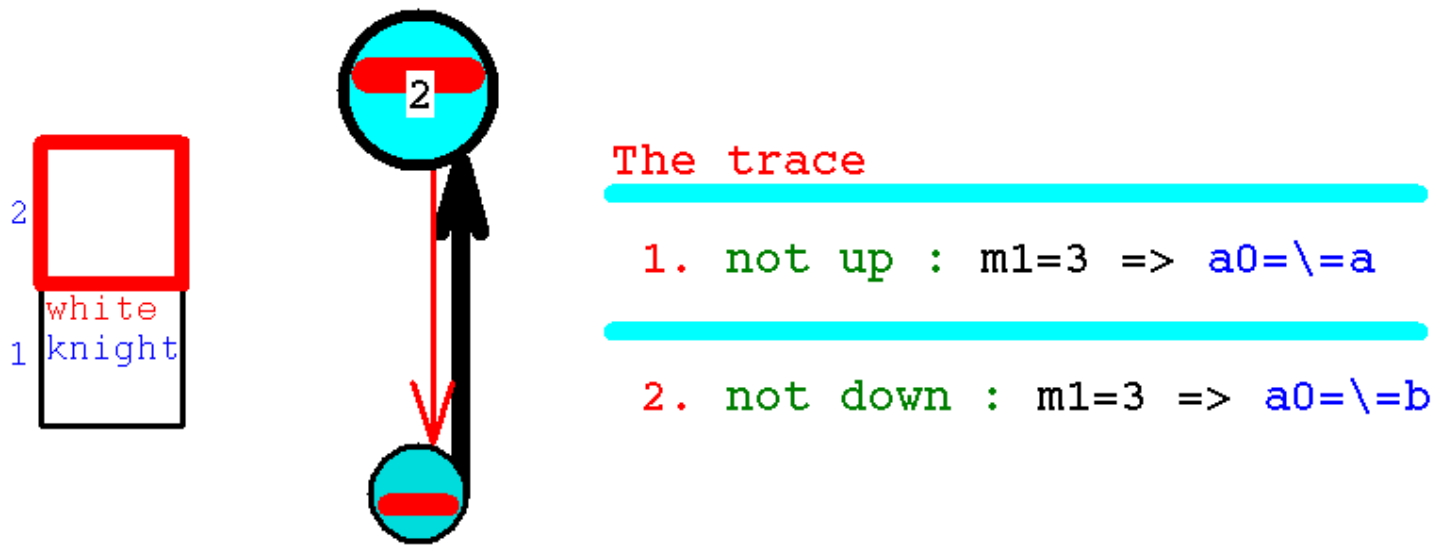

Figure 9

We said that the language for description of worlds will tell us which the current state of the world is. Where is this state stored? At two locations – first, the current state of each ED model and second, the moving traces. For example, in Figure 8 we can see how the moving trace presents the position on the chessboard.

If the language for description of worlds were a standard programming language, its memory would hold the values of the variables and of the arrays. By analogy, we can say that the current state of the ED model is the value of one variable and the value of one moving trace is the value of one array.

The value of the current state of an ED model is usually a number when the model is deterministic or several numbers if the ED model has several current states (the value can be a *belief* if different states have different probabilities). The value of each cell of the moving trace array will consist of several numbers because one state can have many moving traces. Certainly, the permanent traces can also be more than one.

## 10. Algorithms

Now that we have described the basic rules of the chess game and the position of the chessboard, the next step is to describe how the chess pieces move. For this purpose we will resort to the concept of algorithm.

Most papers do not distinguish between an algorithm and a computable function. This is not correct because the algorithm is a given action while the computable function is the result from that action. We should differentiate actions from results. E.g. making pancakes is not the same as the actual pancake. The result from the execution of an algorithm depends on the specific world in which we execute that algorithm. This means that in a different world the pancake making algorithm may produce a different result. That result can be, for example, a computable function or a spacecraft.

### 10.1 What is an algorithm

For most people an algorithm is a Turing machine. The reason is that they only look at $\mathbb{N} \rightarrow \mathbb{N}$ functions and see the algorithm as something which computes these functions. To us, an algorithm will describe a sequence of actions in an arbitrary world. In our understanding, algorithms include cooking recipes, dancing steps, catching a ball and so forth. We just said a sequence of actions. Let us put it better and change this to a sequence of events. An action is an event, but not every event is an action or at least our action – it can be the action of another agent. The description of the algorithm will include our actions as well as other events. For example, we wait for the water in our cooking pan to boil up. The boiling of water is an event which is not our action.

In our definition, an algorithm can be executed without our participation at all. The Moonlight Sonata, for example, is an algorithm which we can execute by playing it. However, if somebody else plays the Moonlight Sonata, it will still be an algorithm albeit executed by someone else. When we hear the piece and recognize that it is the Moonlight Sonata, we would have recognized the algorithm even though we do not execute it ourselves.

Who actually executes the algorithm will not be a very important issue. It makes sense to have somebody demonstrate the algorithm to us first before we execute it on our own.

We will consider three versions of algorithm:

1. Railway track;

2. Mountain footpath;

3. Going home.

In the first version there will be restrictions which do not allow us to deviate from the execution of the algorithm. For example, when we board a coach, all we can do is travel the route. We cannot make detours because someone else is driving the coach. Similarly, when listening to someone else's performance of the Moonlight Sonata, we are unable to change anything because we are not playing it.

In the second version, we are allowed to make detours but then consequences will occur. A mountain footpath passes near an abyss. If we go astray of the footpath we will fall in the abyss.

In the third version, we can detour from the road. After the detour we can go back to the road or take another road. The Going home algorithm tells us that if we execute it properly, we will end up at our home, but we are not anyhow bound to execute it or execute in exactly the same way.

Typically, we associate algorithms with determinacy. We picture in our mind a computer program where the next action is perfectly known. However, even computer programs are not single-threaded anymore. With multi-threaded programs it is not very clear what the next action will be. Cooking recipes are even a better example. When making pancakes, we are not told which ingredient to put first – eggs or milk. In both cases we will be executing the same algorithm.

Imagine an algorithm as a walk in a cave. You can go forward, but you can also turn around and go backward. The gallery has branches and you are free to choose which branch to take. Only when you exit the cave you will have ended the execution of the cave walk algorithm. In other words, we imagine the algorithm as a directed graph with multiple branches and not as a road without any furcations.

## 10.2 The algorithm of chess pieces

We will use algorithms to describe the movement of chess pieces. We will choose the Railway track version (the first one of the versions examined above). This means that when you lift a piece you will invoke an algorithm which prevents you from making an incorrect move.

We could have chosen the Mountain footpath which allows you to detour from the algorithm, but with consequences. For example, lift the piece and continue with the algorithm, but if you break it at some point the piece will escape and go back to its original square.

We could have chosen the Going home version where you can move as you like but can drop the piece only at places where the algorithm would put it if it were properly executed. That is, you have full freedom of movement while the algorithm will tell you which moves are the correct ones.

We will choose the first version of the algorithm mainly because we have let the agent play randomly and if we do not put her in some rail track, she will struggle a lot in order to make a correct move. Moreover, we should consider how the agent would understand the world. How would she discover these algorithms? If we put her on a rail track, she will learn the algorithm – like it or not – but if we let her loose she would have hard time trying to guess what the rules for movement are. For example, if you demonstrate to a school student the algorithm of finding a square root, he will learn to do so relatively easily. But, the kid's life would become very difficult if you just explain to him what is square root and tell him to find the algorithm which computes square roots. You can show the student what a square root is with a definition or examples, but he would grasp the algorithm more readily if you demonstrate hands-on how it works.

What will be the gist of our algorithms? These will be Event-Driven models. There will be some input event which triggers the algorithm and another output event which will put an end to the execution of the algorithm. Later on we will clone the outputs in two (successful and unsuccessful output).

Each chess piece will have its algorithm:

## 10.3 The king and knight algorithms

The king's algorithm (Figure 10) will be the simplest one. The input event will be *king lifted*. The input point will be state 1 (this applies to all algorithms described here). The events will be four (*left*, *right*, *forward* and *backward*).

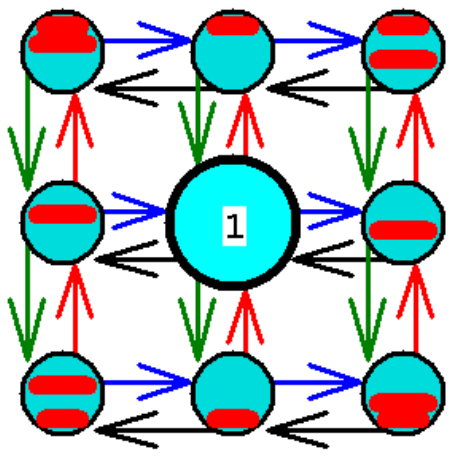

Figure 10

The trace will consist of four events (*left move not allowed*, *right move not allowed*, etc.). These four events (traces) will restrict the king's movement to nine squares. Thus, the four events (traces) will be the rail tracks in which we will enter and which will not let us leave the nine squares until we execute the algorithm. In Figure 10, the four traces are marked with red horizontal lines. For example, the three upper states have the first trace which means that the king cannot move forward from these three states.

We may drop the lifted piece (the king) whenever we wish. Certainly, there will be other restrictive rules and algorithms. E.g. *we cannot capture our own pieces* is an example of other restrictions, which however are not imposed by this algorithm. If we drop the piece in state 1, the move will not be real but fake. If we drop the piece in another state, then we would have played a real move.

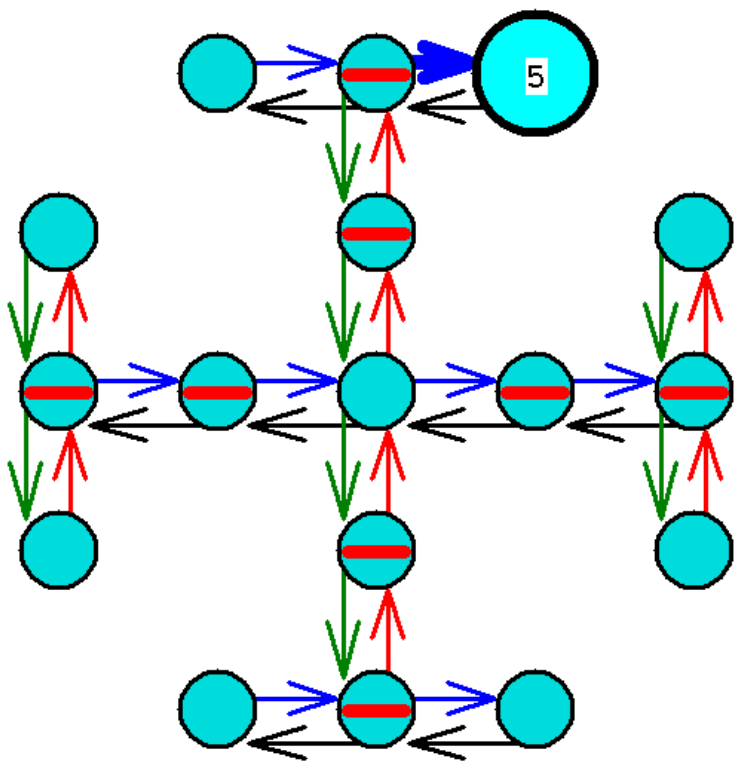

Figure 11

The knight's algorithm is somewhat more complicated (Figure 11). The main difference with the king's algorithm is that here we have one more trace. This trace restricts us such that in certain states we cannot drop the lifted piece. (Only this trace is marked in Figure 11, the other four traces are not.) In this algorithm we have only two options – play a correct move with the knight or play a fake move by returning the knight to the square which we lifted it from.

## 10.4 The rook and bishop algorithms

Although with less states, the rook's algorithm is more complex (Figure 12). The reason is that this algorithm is non-deterministic. In state 3 for example there two arrows for the *move forward* event. Therefore, two states are candidates to be the next state. This non-determinacy is resolved immediately because in state 1 it must be seen that a piece has been lifted from that square while the opposite must be seen in state 3. Therefore, we have a trace which resolves the algorithm's non-determinacy immediately.

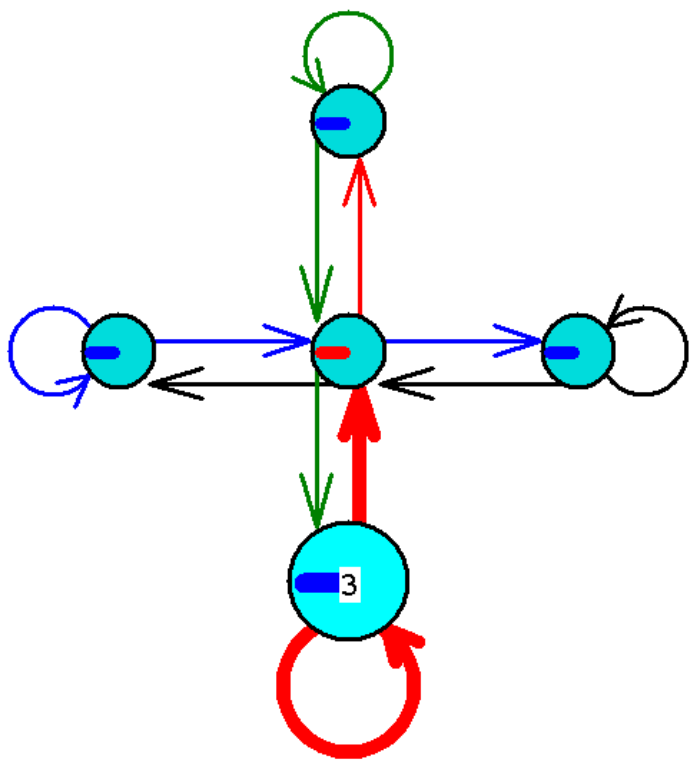

Figure 12

Even more complicated is the bishop's algorithm (Figure 13). The reason is that we cannot move the bishop diagonally outright and have to do this in two steps: first a horizontal move and then a vertical move. If the event *left* occurs in state 1, we cannot know whether our diagonal move is *left and forward* or *left and backward*. This is another non-determinacy which cannot however be resolved immediately. Nevertheless, the non-determinacy will be resolved when a *forward* or *backward* event occurs. In these two possible states we have traces which tell us "*forward move not allowed"* in state 8 and "*backward move not allowed"* in state 2. If the *no-forward* restriction applied in both states, the *forward* event would breach the algorithm. But in this case the event is allowed in one of the states and disallowed in the other state. So, the *forward* event is allowed, but if it occurs state 8 will become inactive and the non-determinacy will be resolved. (In Figure 13 we have marked only the *no-forward* and the *no-backward traces*.)

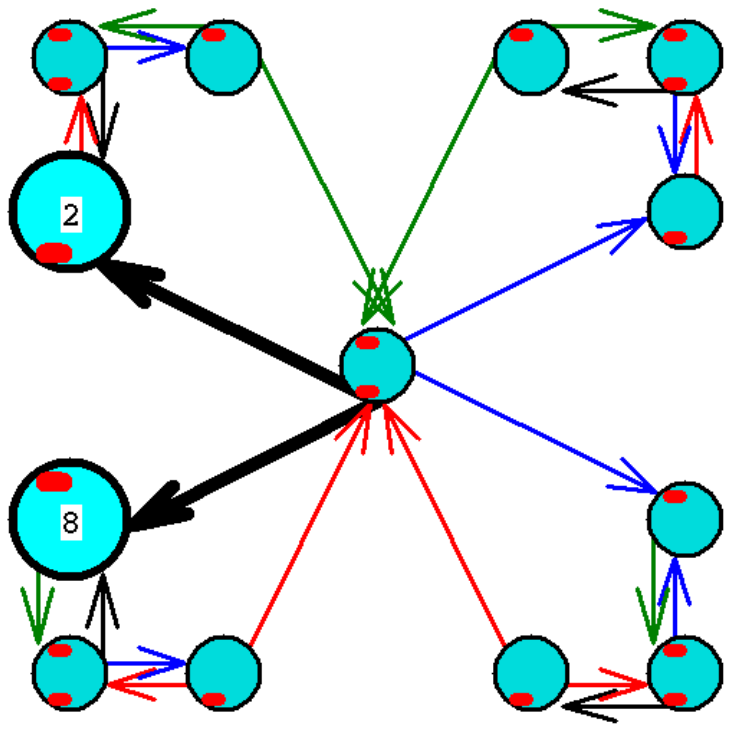

Figure 13

The most complicated algorithm is that of the queen because it is a combination of the rook and bishop algorithms. The pawn's algorithm is not complicated, but in fact we have four algorithms: for white/black pawns and for moved/unmoved pawns.

### 10.5 The Turing machine

So far we described the chess pieces algorithms as Event-Driven (ED) models. Should we assert that each algorithm can be presented as an ED model? Can the Turing machine be presented in this way?

We will describe a world which represents the Turing machine. The first thing we need to describe in this world is the infinite tape. In the chess game, we described the chessboard as the moving trace of some ED model with 64 states. Here we will also use a moving trace, however we will need a model with countably many states. Let us take the model in Figure 6. This is a model of a tape comprised of eight cells. We need the same model which has again two events (*left* and *right*), but is not limited to a *leftmost* and *rightmost* state. This means an ED model with infinitely many states. So far we have only used models with finitely many states.

Now we will have to add some infinite ED models which nevertheless have structures as simple as this one. In this case the model is merely a counter, which keeps an integer number (i.e. an element of $\mathbb{Z}$). The counter will have two operations (*minus one* and *plus one*) or (*move left* and *move right*). The addition of an infinite counter expands the language for description of the world, but as we said we will keep expanding the language in order to cover the worlds we aim to describe.

What kind of memory will this world have? We must memorize the counter value (that is the cell on which the head of machine is placed). This is an integer number. Besides this, we will need to memorize what is stored on the tape. For this purpose we will need an infinite sequence of 0 and 1 numbers, which is equinumerous to the continuum. We usually use Turing machines in order to compute $\mathbb{N} \rightarrow \mathbb{N}$ functions. In this case we can live only with configurations which use only a finite portion of the tape, i.e. we can consider only a countable number of configurations, however, all possible configurations of the tape are continuum many.

**Note:** The agent's idea of the state of the world will be countable even though the memory of the world is a continuum. In other words, the agent cannot figure out all possible configurations on the tape, but only a countable subset of these configurations. In this statement we imagine the agent as an abstract machine with an infinite memory. If we image the agent as a real computer with a finite memory, in the above statement we must replace *countable* with *finite*. Anyway, if the agent is a program for a real computer, the finite memory would be enormous, so for the sake of simplicity we will deem it as countable.

Thus, we have described the tape of the Turing machine with an infinite ED model. In order to describe the head of the machine (the algorithm *per se*) we will need another ED model. We will employ the Turing machine in order to construct the second ED model.

We assumed that the machine uses two letters (0 and 1). Let us construct an ED model with four events:

- *write(0),*
- *write(1),*
- *move left,*
- *move right.*

Then each command to the machine will be in the following format:

- if observe(0) then write Symbol_0, move Direction_0, goto Command_0
- if observe(1) then write Symbol_1, move Direction_1, goto Command_1

Here Symbol_i, Direction_i and Command_i have been replaced with concrete values. For example:

- if observe(0) then write(1), move left, goto $s_3$
- if observe(1) then write(0), move right, goto $s_7$

We will replace each command with four states which describe it. The above command will take the following form:

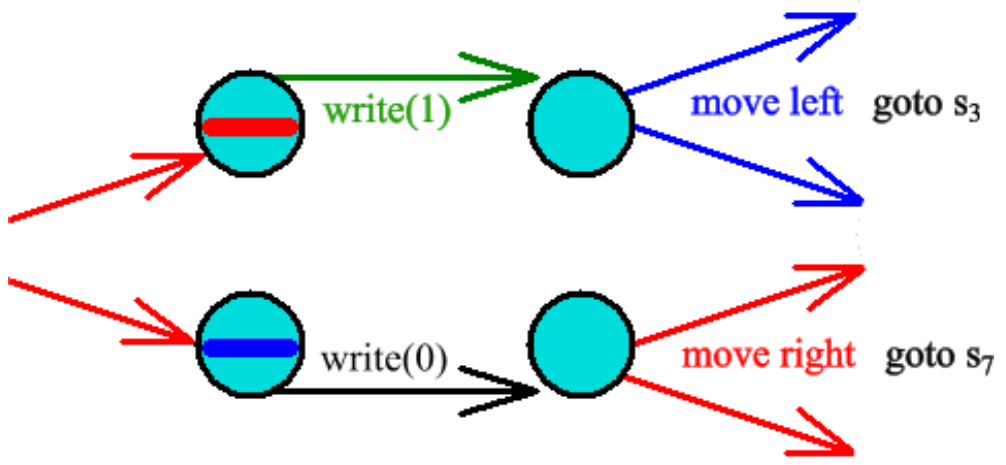

Figure 14

In Figure 14, the input is over the *move right* event. In fact there will be many input paths – sometimes over the *move left* event and other times over the *move right* event. Importantly, the input will be non-deterministic but the non-determinacy will be resolved immediately because the first two states have a trace. In the top state the event "observe(0)" must always occur and in the bottom state the "observe(0)" event must never occur.

Thus, each state of the machine is replaced with four states as shown in Figure 14 and then the individual quaternaries are interconnected. For example, the quaternary in Figure 14 connects to the quaternary in $s_3$ by arrows over the *move left* event and to the quaternary in $s_7$ by arrows over the *move right* event.

We have to add some more trace to accommodate the rule that only one of the four events is possible in each state. The new trace should tell us that the other three events are impossible. We should do this in case we want a Railway track type of algorithm. If we prefer a Mountain footpath algorithm, we must add a trace which tells which consequences will occur if one of the other three events happens. If we wish a Go home algorithm, then the other three events must lead to a termination of the algorithm.

Thus we presented the Turing machine through an Event-Driven model or, more precisely, through two ED models – the first one with infinitely many states and the second one with a finite number of states (the number of machine states multiplied by 4).

Who executes the algorithm of the Turing machine? We may assume the four events are actions of the agent and that she is the one who runs the algorithm. We may also assume that the events are acts of another agent or that the events just happen. In that case the agent will not be the executor of the algorithm, but just an observer. In the general case, some events in the algorithm will be driven by the agent and all the rest will not. For example, "I pour water in the pot" is an action of the agent while "The water boils up" is not hers action. The agent can influence even those events which are not driven by hers actions. This is described in Dobrev (2019b). For these events the agent may have some "preference" and by hers "preferences" the agent could have some influence on whether an event will or will not occur.

## 10.6 Related work

Importantly, this paper defines the term *algorithm* as such. Very few people bother to ask what is an algorithm in the first place. The only attempts at a definition I am aware of are those Moschovakis (2001; 2018). In these works Moschovakis says that most authors define algorithms through some abstract machine and equate algorithms with the programs of that abstract machine. Moschovakis goes on to explain what kind of an algorithm definition we need – a generic concept which does not depend on a particular abstract machine. The computable function is such a concept, but for Moschovakis it is too general so he seeks to narrow it down to a more specific concept which reflects the notion that a computable function can be computed by a variety of substantially different algorithms. This is a tall aim which Moschovakis could not reach in Moschovakis (2001). What he did there can be regarded as a new abstract machine. Indeed, the machine is very interesting and more abstract than most known machines, but again we run into the trap that the machine's program may become needlessly complicated and in this way morph into a new program which implements the same algorithm. Although Moschovakis (2001) does not achieve the objective of creating a generic definition of an algorithm, Moschovakis himself admits that his primary objective is to put the question on the table even if he may not be able to answer it. His exact words are: “my chief goal is to convince the reader that the problem of founding the theory of algorithms is important, and that it is ripe for solution.“

## 11. Objects

### 11.1 Properties

Having defined the term *algorithm*, we will try to define another fundamental concept: *property*. For the definition of this concept we will again resort to the Event-Driven models. A property is the phenomenon we see when we observe an object which possesses that property. Phenomena are patterns which are not observed all the times but only from time to time. Given that the other patterns are presented through ED models, it makes perfect sense to present properties through ED models, too.

The difference between a pattern and a property will be that the pattern will be active on a permanent basis (will be observed all the time) while the property will be observed from time to time (when we observe the object which possesses that property).

### 11.2 What is an object?

The basic term will be *property* while *object* will be an abstraction of higher order. For example, if in the chess world one observes the properties *white* and *knight* he may conclude that there is a *white knight* object which is observed and which possesses these two properties. We may dispense of objects and simply imagine that some properties come and go, i.e. some phenomena appear and disappear. However, object abstraction is mandatory for understanding complex worlds.

### 11.3 The second coding

The agent’s output consists of four characters only which made us use coding in other to describe the eight possible actions of the agent. The input is also limited to four characters. While it is true that the input will depict to us only one square rather than the full chessboard, four characters are still too little because a square can accommodate six different pieces in two distinct colors. Furthermore, we need to know whether the pawn on the square has moved and whether the lifted piece comes from that square. How can one present all that amount of information with four characters only?

That information may not necessarily come to the agent for one step only. The agent can spend some time staying on the square and observing the input. As the agent observes the square, she may spot various patterns. The presence or absence of each of these patters will be the information which the agent will receive for the square she observes. Although the input characters are only four, the patterns that can be described with four characters are countless.

Let us call these patterns *properties* and assume that the agent is able to identify (capture) these patterns. We will further assume that the agent can capture two or more patterns even if they are layered on top of each other. Thus, the agent should be able to capture the properties *white* and *knight* even when these properties appear at the same time.

How would the properties look like? In the case of chess pieces, the patterns and algorithms of their movement are written by a human who has an idea of the chess rules and of how the pieces move. The properties are not written by a human and are generated automatically. As an example, Figure 15 depicts the property *king*. That property appears rather bizarre and illogical. The reason is, as we said, that the property is generated automatically in a random way. It is not written by us because we do not know how the king would look like. How the king looks like does not matter. What matters is that the king should have a certain appearance such that it can be recognized by the agent. In other words, the king should have some face, but how that face would look like is irrelevant.

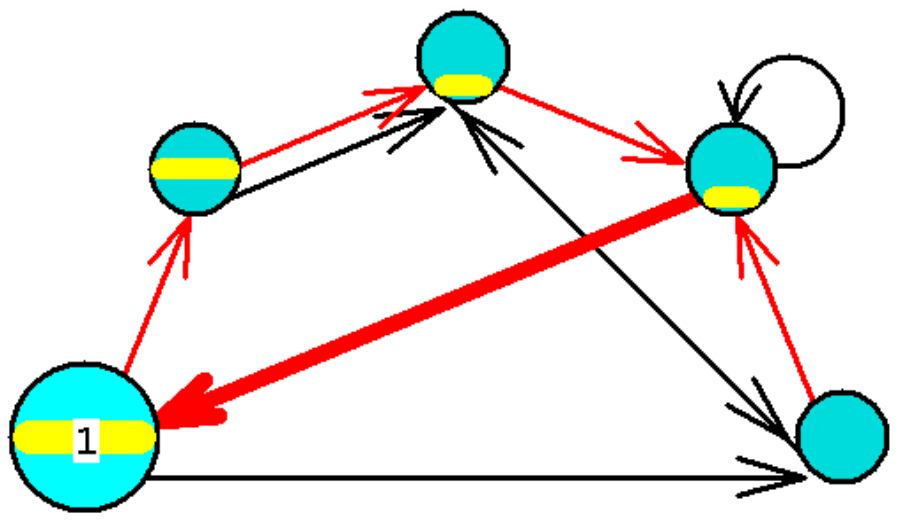

Figure 15

In our program (Dobrev, 2020a) there are 10 properties and each of them has some trace. When several properties are active at the same time, each of them has an impact (via its trace) on the agent's input. Sometimes these impacts may be contradictory. For example, one property tells us that the next input must be the character ***x***, while another property insists on the opposite (the input must not be ***x***). The issue at hand is then solved by voting. The world counts the votes for each decision and selects the one which reaps the highest number of votes. As concerns contradictory recommendations, they will cancel each other.

### 12. A two-players chess game

We will sophisticate the chess game by adding another agent: the opponent (antagonist) who will play the black pieces. This will induce non-determinacy because we will not be able to tell what move the opponent will play. Even if the opponent is deterministic, hers determinacy would be too complicated for us to describe it.

### 12.1 A deterministic world

So far we have described a chess world where the agent plays solitaire against herself. We have written a program (Dobrev, 2020a) which contains a simple description of this world and by this description emulates the world. Run this program and see how simple that description happens to be. The description consists of 24 modules presented in the form of Event-Driven models (these models are directed graphs with a dozen states each). The ED models of the chess game belong to three types: (5 patterns + 9 algorithms + 10 properties = 24 models). In addition to the ED models there are also two moving traces (i.e. two arrays). We added also six simple rules which we need as well. These rules provide us with additional information about how the world has changed. For example, when we lift a piece, the property *Lifted* will appear in the square of that piece. This rule looks like this:

$$up,\ here \Rightarrow copy(Lifted)$$

(If we lift a piece and if we are in the square <*X, Y*>, then the property *Lifted* will replace all properties which are in this square at the moment.)

We are able to formulate these rules owing to the fact that we have the context of the chessboard (the moving trace from Figure 8). If we had no idea that a chessboard exists, we would not be able to formulate rules for the behavior of the pieces on that board. In the demonstration program (Dobrev, 2020a) the agent plays randomly. Of course the agent's actions are not important. What matters is the world and that we have described it.

The description thus obtained is deterministic, i.e. the initial state is determined and every next state is determined. A deterministic description means that the described world is free from randomness. Should the description of the world be deterministic? Should we deal only with deterministic descriptions? The idea that the world may be deterministic seems outlandish. And even if it were, we need not constrain ourselves to deterministic descriptions.

If we apply a deterministic description to a non-deterministic world, that description will very soon exhibit its imperfections. Conversely, the world may be deterministic but its determinacy may be too complicated and therefore beyond our understanding (rendering us unable to describe it). Accordingly, instead of a deterministic description of the world will find a non-deterministic description which works sufficiently well.

Typically, the world is non-deterministic. When we shoot at a target may miss it. This means that our actions may not necessarily yield a result or may yield different results at different times.

We will assume that the model may be non-deterministic. For most authors, non-deterministic implies that for each possible event there is one precisely defined probability. In Dobrev (2018; 2019b) we showed that the latter statement is too deterministic. Telling the exact probability of occurrence for each and every event would be an exaggerated requirement. Accordingly, we will assume that we do not know the exact probability, but only the interval [*a, b*] in which this probability resides. Typically that would be the interval [*0, 1*] which means that we are in total darkness as regards the probability of the event to occur.

## 12.2 Impossible events

We said the world would be more interesting if we do not pay against ourselves but against another agent who moves the black pieces.

For this purpose we will modify the fifth ED model (the one which tells us which pieces we are playing with – white or black). This model has two states which are switched by the event *change*. Previously we defined that event as "real_move" (this is the event when we make a real move while in "fake_move" we only touch a piece). We will change the definition of that event and define it as *never* (this is the opposite of *every time*). This change produces a world where the agent cannot switch sides.

Does it make sense to describe in our model events which can never happen? The answer is yes, because these events may happen in our imagination. I.e. even though these events not occur, we need them in order to understand the world. For example, we are unable to fly or change our gender, but we can do this in our imagination. The example is not very appropriate, because we can already fly and change our gender if we wish to. In other words, we can imagine impossible events in our minds. Also, at some point, these impossible events may become possible.

We will use the impossible event *change* in order to add the rule that we cannot play a move after which we will be in check (our opponent will be able to capture our king). Figure 16 depicts the algorithm which describes how we switch sides (turn the chessboard around) and capture the king. If an execution of that algorithm exists, then the move is incorrect. (Even if an execution exists the algorithm cannot be executed, because it includes an impossible event.)

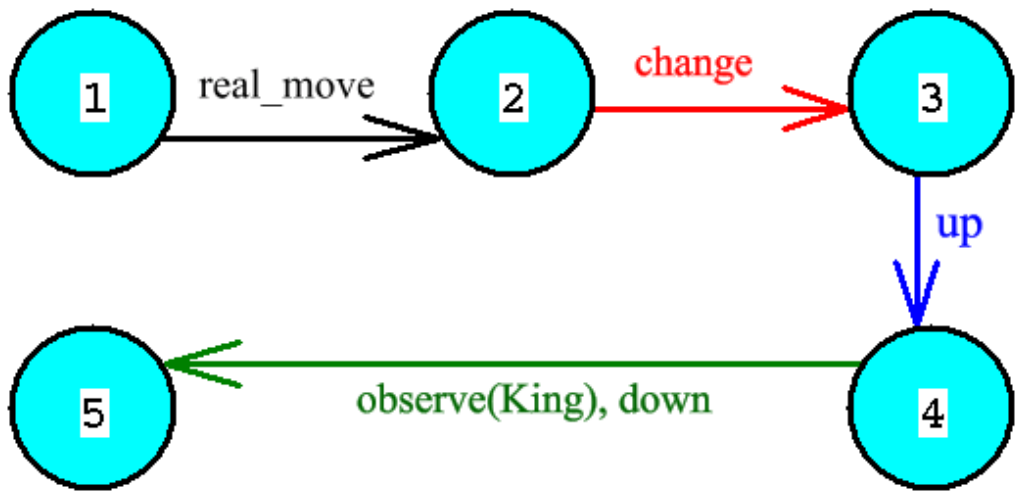

Figure 16

This algorithm is more consistent with our understanding of algorithms. While the algorithms of the chess pieces were directed graphs with multiple branches, this one is a path without any branches. In other words, this algorithm is simply a sequence of actions without any diversions.

This algorithm needs some more restrictions (traces) which are not shown in Figure 16. In state 1 for example we cannot move in any of the four directions (otherwise we could go to another square and play another move). The *change* event can only occur in state 2 and not in any other state. In state 4 we have the restriction “not observe(King) ⇒ not down“ which means that the only move we can make is to capture the king.

Part of this algorithm is the impossible action *change*. As mentioned above, although this action is not possible, we can perform it in our imagination. This event may be part of the definition of algorithms which will not be executed but are still important because we need to know whether their executions exist.

**Note:** In this paper, by saying that an algorithm can be executed we mean that it can be executed successfully. This means that the execution may finish in a final (accepting) state or with an output event (successful exit).

### 12.3 The second agent

The algorithm in Figure 16 would become simpler if we allow the existence of a second agent. Instead of switching between the color of the pieces (turning the chessboard around) we will replace the agent with someone who always plays the black pieces. Thus, we will end up with an algorithm performed by more than one agent, which is fine because these algorithms are natural. For example, “I gave some money to someone and he bought something with my money” is an example of an algorithm executed by two agents.

The important aspect here is that once we move a white piece, we will have somebody else (another agent) move a black piece. While in the solitaire version of the game we wanted to know whether a certain algorithm is possible, in the two players version we want a certain algorithm to be actually executed. Knowing that a certain algorithm is possible and the actual execution of that algorithm are two different things. Knowing that “someone can cook pancakes” is okay but “your roommate cooking pancakes this morning” is something different. In the first case you will know something about the world while in the latter case you will have some pancakes for breakfast. If the actual execution of an algorithm will be in the hands of agent, then it does matter who the executing agent is. E.g. we will suppose that the pancakes coming from the your roommate’s hands will be better than those cooked by you.

We will assume that after each “real_move” we play, the black-pieces agent will execute the algorithm in Figure 17.

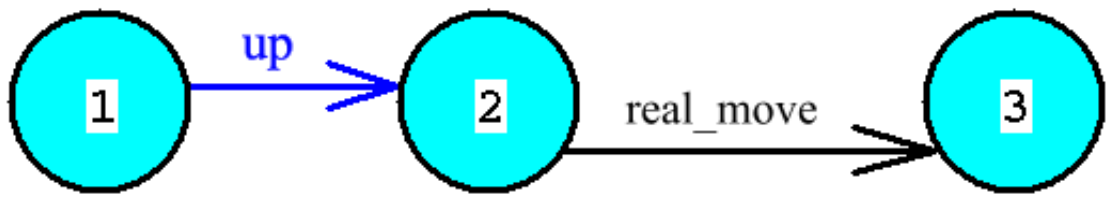

Figure 17

The execution of an algorithm does not happen outright because it is a multi-step process. Nevertheless, we will assume here that the opponent will play the black pieces right away (in one step). When people expect someone to do something, they tend to imagine the final result and ignore the fact that the process takes some time. Imagine that “Today is my birthday and my roommate will cook pancakes for me”. In this reflection you take the pancakes for granted and do not bother that cooking the pancakes takes some time.

As mentioned already, it matters a lot who the black-pieces agent is. Highly important is whether the agent is friend or foe (will she assist us or try to disrupt us). The agent’s smartness is also important (because she may intend certain things but may not be smart enough to do these things). It is also important to know what the agent can see. In the chess game we assume that the agent can see everything (the whole chessboard) but in other worlds the agent may only be able to know and see some part of the information. The agent's location can also be important. Here will assume that it is not important, i.e. wherever the agent is, she may move to any square and lift the piece in that square. In another assumption the agent's position may matter because the pieces that are nearer the agent may be more likely to be moved than the more distant pieces.

## 12.4 Agent-specific state

We assumed here that the second agent sees a distinct state of the world. I.e. the second agent has hers distinct position $<x, y>$ on the chessboard (the square which she observes). We also assume that the second agent plays with black pieces as opposed to the protagonist playing with the white pieces.

We assume that the two agents change the world according to the same model but the memory of that model (the state of the world) is specific to each agent. We might assume that the two states of the world have nothing in common, but then the antagonist's actions will not have any impact on the protagonist's world. Therefore, we will assume that the chessboard position is the same for both agents (i.e. the trace in Figure 8 is the same for both). We will further assume that each agent has hers own coordinates and a specific color for hers pieces (i.e. the active states of Event-Driven models 2, 3 and 5 are different for the two agents). As concerns the other ED models and the trace in Figure 9, we will assume that they are also specific to each agent, although nothing prevents us from making the opposite assumption.

Had we assumed that the two agents share the same state of the world, the algorithm in Figure 17 would become heavily complicated. The antagonist would first turn the chessboard around (*change*), then play hers move and then turn the chessboard around again in order to leave the protagonist's world unchanged. Moreover, the antagonist would need to go back to hers starting coordinates $<x, y>$ (these are the protagonist's coordinates). It would be bizarre to think that the separate agents are absolutely identical and share the same location. The natural way of thinking is that the agents are distinct and have distinct, but partially overlapping states of the world. For example, "Right now I am cooking pancakes and my roommate is cooking pancakes, too". We may be cooking the same pancakes or it may be that my pancakes have nothing to with his.

**Note:** It is not very accurate to say that the world has distinct states for both agents. The world is one and it has only one state. It would be more accurate to say that we have changed the world and now we have a world with more complicated states. Let the new set of states be $S''$. We can assume that $S'' = S \times S$. The questions that are common to the two agents have remained the same, but the other questions are now bifurcated. For example, "Where am I?" is replaced with the questions "Where is the protagonist?" and "Where is the antagonist?". Thus, from the model where the states are $S$ we have derived a new model where the states are $S''$. The difference between $S$ and $S''$ is that the states in $S$ describe the state of one agent (without telling us which is that agent), while the states in $S''$ describe the states of the two agents. (In both cases the overall state of the world is described as well.) The new model describes (i) the world through the two agents and (ii) how the agents change their states according to the first model. Nevertheless, it is more natural to assume that the world has different states for the two agents and these agents change their states according to the first model which operates only with the questions that apply only to one of the agents.

## 12.5 Non-computable rule

So far we have described the first world in which the agent plays solitaire against herself and have written the program (Dobrev, 2020a) which emulates the first world. The program (Dobrev, 2020a) is a model which describes the first world. We have also described a second world in which the agent plays against some opponent (antagonist). Now, can we also create a program which emulates the second world?

In the second world we added a statement which says “This algorithm can be executed”. (This statement was to be added in the first world, because playing moves after which we are in check is not allowed in the first world, too. For the time being the program (Dobrev, 2020a) allows us to make this kind of moves.) In the second world we also added the operation “Opponent executes an algorithm”. In the general case that statement and that operation are undecidable (more precisely, they are semi-decidable).

For example, let us take the statement “This algorithm can be executed”. In this particular case the question is whether the opponent can capture our king, which is fully decidable because the chessboard is finite, has finitely many positions and all algorithms operating over the chessboard are decidable. In the general case the algorithm may be a Turing machine and then the statement will be equivalent to a halting problem.

The same can be said of the operation “Opponent executes an algorithm”. While the algorithm can be executed by many different methods, the problem of finding at least one of these methods is semi-decidable. In the particular case of the chess game we can easily find one method of executing the algorithm, or even all methods (i.e. all possible moves). In the general case, however, the problem is semi-decidable.

Therefore, in this particular case we can write a program which emulates the second world, provided however that we have to select the opponent’s behavior because it can go in many different paths. In other words, in order to create a program which emulates the chess world, we should embed in it a program which emulates a chess player.

In the general case however, we will not be able to write a program which emulates the world we have described. Thus, the language for description of worlds is already capable of describing worlds that cannot be emulated by a computer program. We said in the very beginning that the model may turn out to be non-computable. Writing a program which computes non-computable model is certainly impossible.

However, being unable to write a program that emulates the world we have described is not a big issue, because we are not aiming to emulate the world, but write an AI program which acts on its understanding of the world (the description of the world which it has found) in order to successfully plan its future moves. Certainly, the AI program can proceed with one emulation of the world, play out some of its possible future developments and select the best development. (Essentially this is how the Min-Max algorithm of chess programs works.) I.e. making an emulation of the world would be a welcome though not mandatory achievement.

Besides being unable to produce a complete emulation of the world (when the model is non-computable), AI would be unable to even figure out the current state of the world (when the possible states are continuum many). Nevertheless, AI will be able to produce a partial emulation and figure out the state of the world to some extent. For example, if there is an infinite tape in the world and this tape carries an infinite amount of information, AI will not be able to discern the current state of the world, but would be capable of describing some finite section of the tape and the information on that finite section.

Even the Min-Max algorithm is not a complete emulation due to combinatorial explosion. Instead, Min-Max produces partial emulation by only traversing the first few moves. If the description of the world contains a semi-decidable rule, AI will use that rule only in one direction. An example is the rule which says that “A statement is true if there is proof for that statement”. People use that rule if (i) a proof exists and (ii) they have found that proof. If a proof does not exist, the rule is not used because we cannot ascertain that there is not any proof at all.

## 13. Agents

Our next abstraction will be the agent. Similar to objects, we will not be able to detect agents outright but will gauge them indirectly by observing their actions. The detection of agents is a difficult task. People manage to detect agents, however, they need to search them everywhere. Whenever something happens, people quickly jump into the explanation than some agent has done that. In peoples’ eyes, behind every event there lurks a perpetrator which can be another human or an animal or some deity. Very seldom they would accept that the event has occurred through its own devices. AI should behave as people do and look for agents everywhere.

Once AI detects an agent it should proceed to investigate the agent and try to connect to it. To detect an agent means to conjure up an agent. When AI conjures up an existing agent, then we can say that AI has detected the agent. When AI conjures up a non-existing agent, the best we can say is that AI has conjured up a non-existing thing. It does not really matter whether agents are real or fictitious as long as the description of the world obtained through these agents is adequate and yields appropriate results.

### 13.1 Interaction between agents

AI will investigate agents and classify them as friends or foes. It will label them as smart or stupid and as grateful or revengeful. AI will try to connect to agents. To this end, AI must first find out what each agent is aiming at and offer that thing to the agent in exchange of getting some benefit for itself. This exchange of benefits is the implementation of a coalition policy. Typically it is assumed that agents meet somewhere outside the world and there they negotiate their coalition policy. But, because there is no such place outside the world, we will assume that agents communicate within the walls of the world. The principle of their communication is: “I will do something good for you and expect you to do something good for me in return”. The other principle is “I will behave predictably and expect you to find out what my behavior is and start implementing a coalition policy (engage in behavior which is beneficial to both of us)”.

This how we communicate with dogs. We give a bone to a dog and right away we make friends. What do we get in return? They will not bite us or bark at us, which is a fair deal. As time goes by the communication may become more sophisticated. We may show an algorithm to the agent and ask her to replicate it. We can teach the dog to “shake hands” with a paw. Further on, we can get to linguistic communication by associating objects with phenomena. For example, a spoken word is a phenomenon and if this phenomenon is associated with a certain object or algorithm, the agent will execute the algorithm as soon as she hears the word. E.g. the dog will come to us as soon as it hears its name or bring our sleepers when it hears us saying “sleepers”.

### 13.2 Signals between agents

When it comes to interaction or negotiation, we need some sort of communication. This takes us to the signals which agents send to their peers. We do not mean pre-arranged signals, but ones which an agent choose to send and the others decipher on the basis of their observations. One example is “Pavlov’s Dog” (Pavlov, 1902). Pavlov is the agent who decides to send a signal by ringing a bell before he feeds the dog. The other agent is the dog which manages to comprehend the signal.

Where an agent sends a signal to another agent, the latter need not necessarily realize that this is a signal and has been send by someone else who is trying to tell her something. In the previous example, Pavlov’s dog has not any idea that Pavlov is the one ringing the bell to signal that lunch is ready. The dog simply associates the *ringing* event to the *feeding* event. Thus, when we send a signal we can remain anonymous. This ultimately means that we can influence another agent without that agent ever realizing that she is under somebody else’s influence.

Another way of sending a signal is to show something (provide some information). For this purpose we need to know what the other agent can see and when. For example, when a dog growls at us, it shows us its fangs. We see that the dog has fangs – something which we know by default – but what we realize in this case is that the dog has decided to remind us of this fact and understand the message as “The dog issues a warning that it may use its fangs against us”.

In addition to natural signals (ones which we can guess ourselves), there may be pre-established signals. Let us have a group of agents who have already established some signals between them. When a new agent appears, she may learn a signal from one of the agents and then use it in hers communication with other agents. An example for such signals are the words in our natural language. We learn the words from an agent (e.g. from our mothers) and then use the same words in order to communicate with other agents.

### 13.3 Exchange of information

When agents communicate, they can exchange information in order to coordinate their actions or to negotiate. An example of information exchange is when an agent shares some algorithm with another agent. The algorithm can be described in a natural language, i.e. it can be presented as a sequence of signals (words) where each signal is associated with an object, phenomenon or algorithm. For example, when we tell somebody how to get to a shop, we explain this algorithm by using words. When we say “Open the door” we rely that the other agent will associate the word “door” with the object *door* and the word “open” with the algorithm *open*. In other words, we rely that the agent knows these words and has an idea of the objects associated with these words.

If we assume that the agent keeps the algorithms in hers memory in the form of Event-Driven models, then the agent should be able to construct an ED model from a description expressed in a natural language and vice versa – describe some ED model in natural language (as long as the agent knows the necessary words).

### 13.4 Communication interface

When creating the AI’s world, we need to equip it with some communication interface to enable it communicate with other agents.

For example, when building a self-driving vehicle, we must give that vehicle some face so that it can communicate with pedestrians and other drivers. Indeed, vehicles have horns and blinkers, but this is not sufficient for full communication. It would be a good idea to add some screen which expresses various emotions. Smiling and winking will be very useful functions.

We usually try see where the other driver is looking at because it is very import for us to know that the other driver has seen us. If this face (screen) is able to turn to our side, it would be an indication that we have been seen.

### 13.5 Related work

Many authors deal with the interaction between agents. Their papers however do not tell us how AI will discover the agents as they assume that the agents have already been discovered and all we need to do is set out rules for reasonable interaction. Goranko, Kuusisto and Rönnholm (2020) for example looks at the case where all agents are friends and their smartness is unlimited. Agents in Goranko et al. (2020) communicate on the basis that they can foretell what the others would do (relying on the assumption that all agents are friends and are smart enough to figure out what would be beneficial for everybody). The most interesting aspect in Goranko et al. (2020) is that the paper raises the issue about the hierarchy between agents (who is more important) and about their hastiness (how patient is each agent). These are principles which real people apply in the real world and it therefore makes sense for AI to also use these principles.

Interaction between agents is as complicated as interaction between humans. As an example, agents in Mell, Lucas, Mozgai and Gratch (2020) negotiate and can even cheat each other.

Gurov, Goranko and Lundberg (2021) as well as the present paper deals with a multi-agent system where the agents do not see everything (Partial Observability). The main difference between Gurov et al. (2021) and this paper is that in Gurov et al. (2021) the world is given (is described by one relation) while in this paper the world is not given and is exactly the thing that has to be found.

## 14. Future work

In this paper we provided a manual description of a world (the chess game) and created the computer program Dobrev (2020a) which emulates the world on the basis of the manual description. Our next problem is the inverse one, namely create a program which should automatically arrive at the same description of the world as the one which we have described manually. That program will use the emulation of the world (Dobrev, 2020a), thanks to it will “live” inside the world and will have to understand it (i.e. to describe it).

In this case we might be tempted to play with marked cards because the program we make has to find a thing, and we know in advance what this thing is. Of course we should not yield to this temptation because we will end up with a program which is able to understand only and exclusively the particular world. It would be much better if our program is able to understand (describe) any world. This however is a tall order because essentially it asks us to build AI. Accordingly, instead of aiming at a program which is capable to understand any world, we would be happy with a program which can understand the given world (Dobrev, 2020a) and the worlds that are proximal to it. The larger the class of worlds our program is able to understand, the smarter that program will be.

## 15. Conclusion

Our task is to understand the world. This means we have to describe it but before we can do so we need to develop a specific language for description of worlds.

We have reduced the task of creating AI to a purely logical problem. Now we have to create a language for description of worlds, which will be a logical language because it would enable the description of non-computable functions. If a language enables only the description of computable functions, it is a programing and not a logical language.

The main building blocks of our new language are Event-Driven models. These are the simple modules which we are going to discover one by one. With these modules we will present patterns, algorithms and phenomena.

We introduced some abstractions. Our first abstraction were objects. We cannot observe the objects directly and instead gauge them by observing their properties. A property is a special phenomenon which transpires when we observe an object which possesses that property. Thus the property is also presented through an ED model.

Then we introduced another abstraction – agents. Similar to objects, we cannot observe agents directly and can only gauge them through their actions.

We created a language for description of worlds. This is not the ultimate language, but only a first version which needs further development. We did not provide a formal description of our language and instead exemplified it by three use cases. That is, instead of describing the language we found the descriptions of three concrete worlds – two versions of the chess game (with one and two agents, respectively) and a world which presents the functioning of the Turing Machine.

**Note:** It is not much of a problem to provide a formal description of a language which covers all the three worlds, but we are aiming elsewhere. The aim is to create a language which can describe any world, and provide a formal description of that language. This is a more difficult problem which we are yet to solve.

We demonstrated that through its simple constituent modules, the language for description of worlds can describe quite complicated worlds with multiple agents and complex relationships among the agents. The superstructure we build on these modules cannot hover in thin air and should rest on some steady fundament. Event-Driven models are exactly the fundament of the language for description of worlds and the base on which we will develop all abstractions of higher order.